\documentclass{article}

\usepackage[utf8]{inputenc}

\usepackage{amsmath}
\DeclareMathOperator*{\argmax}{arg\,max}
\DeclareMathOperator*{\argmin}{arg\,min}

\usepackage{subcaption}
\usepackage{graphicx}
\usepackage{amsfonts}
\usepackage[linesnumbered,ruled,vlined]{algorithm2e}
\usepackage{breqn}
\usepackage{nicematrix}
\usepackage{float}
\usepackage{xcolor}
\usepackage{booktabs}
\usepackage{hyperref}

\usepackage{colortbl}
\usepackage{pifont}
\usepackage{array}
\usepackage{multirow}
\usepackage{caption}
\usepackage{hyperref}

\usepackage{biblatex}
\bibliography{references}

\usepackage[title]{appendix}
\usepackage{etoolbox}
\BeforeBeginEnvironment{appendices}{\clearpage}

\def\email#1{{\tt#1}}

\providecommand{\keywords}[1]
{
  \textbf{Keywords:} #1
}

\title{\bf Classification via score-based generative modelling}
\author{Yongchao Huang \footnote{Author email: \email{\{yongchao.huang@outlook.com\}}}}

\date{June 2022}

\begin{document}

\maketitle

\begin{abstract}
    In this work, we investigated the application of score-based gradient learning in discriminative and generative classification settings. Score function can be used to characterize data distribution as an alternative to density. It can be efficiently learned via score matching, and used to flexibly generate credible samples to enhance discriminative classification quality, to recover density and to build generative classifiers. We analysed the decision theories involving score-based representations, and performed experiments on simulated and real-world datasets, demonstrating its effectiveness in achieving and improving binary classification performance, and robustness to perturbations, particularly in high dimensions and imbalanced situations.
\end{abstract}

\keywords{Score-based modelling, discriminative classification, generative classification, imbalanced learning.}

\section{Generative and discriminative approaches to classification}
\label{Sec:intro}

Generally, there are two approaches to probabilistic classification: generative and discriminative \cite{NIPS2001_Ng}. Assume we have in total $c$ classes, i.e. $y \in \{y_1,y_2,...,y_c\}$, the generative approach models the class-conditional density \footnote{We generally consider a probability mass function (\textit{pmf}) or proabbility density function (\textit{pdf}), denoted by $p(x)$, as density. Either maps a scalar or vector $x$ to a non-negative scalar field, i.e. $p(x): \mathbb{R}^n \rightarrow \mathbb{R}$, with the property of summing (or integrating) to unity.} $p(x|y)$ in some functional or architectural form using data, and computes the class probability via the Bayes rule \cite{GP_Rasmussen}:

\begin{equation} \label{Eq:generative_classification_1}
    p(y_j|x) = \frac{p(x|y_j)p(y_j)}{\sum_{j=1}^{c} p(x|y_j) p(y_j)}    
\end{equation}

\noindent where $j=1,2,...,c$. $p(y|x)$ is also termed the \textit{posterior} probability that an observation $x$ belongs to the $j$-th class, $p(x|y)$ is the generative model, and $p(y_j)$ the prior probability of class $j$, which can be empirically estimated as the fraction of training samples attached to the $j$-th class. Typical generative methods include naive bayes, linear discriminant analysis (LDA) and quadratic discriminant analysis (QDA). Estimating the class-specific density $p(x|y)$ from data is not easy, conventional methods such as \textit{kernel density estimation} (KDE), or modern ones such as generative adversarial networks (GANs) can be used to explicitly or implicitly represent the sampling process, and generate samples from the learned dynamics.

Discriminative approaches, e.g. logistic regression, support vector machine, and decision trees, directly model the posterior $p(y|x)$, or learn a direct map from inputs x to the class labels \cite{NIPS2001_Ng}. A discriminative neural network classifier, for example, may use the \textit{softmax} activation $\sigma(x)$ in its output layer to positively transform and squash feature embeddings $f(x)$ to obtain the probability of an observation $x$ belonging to class $j$:

\begin{equation} \label{Eq:softmax_func}
    p(y_j|x) = \sigma(x) = \frac{e^{f_j(x)}}{\sum_{j=1}^c e^{f_j(x)}}
\end{equation}

\noindent The exponential \textit{response function} (i.e. the inverse of a \textit{link} function, as used in generalized linear models) gives the positive transform. In binary classification (i.e. $c=2$), \textit{softmax} reduces to the \textit{sigmoid} function $ \lambda(x)$:

\begin{equation} \label{Eq:sigmoid_func}
    p(y=1|x) = \lambda(x) = \frac{e^{f(x)}}{1+e^{f(x)}}
\end{equation}

\noindent If $f(x)$ is linear, $\lambda(x)$ is termed the \textit{logistic} response function because it models the \textit{logit} (i.e. log odds ratio) p(y=1)/p(y=0) as a linear combination of features:

\begin{equation}
    f(x) = \omega^T x = \log \frac{p(y=1)}{1-p(y=1)}
\end{equation}

\noindent where $\omega$ hosts the linear coefficients. Another common choice of response function is the \textit{cumulative density function} (CDF) of a standard normal distribution $\Phi(z)=\int_{-\infty}^z \mathcal{N}(x|0,1) dx$, which is termed as \textit{probit} regression. 

There is no certain answer to which approach should be chosen \cite{GP_Rasmussen}. The generative approach provides a principled way to deal with missing values and outliers via access to $p(x)$, and a generative classifier may approach its (higher) asymptotic error faster \cite{NIPS2001_Ng}; discriminative methods are straightforward and gets round the density estimation problem which could be hard in high dimensions due to \textit{curse of dimensionality}, and inaccurate with presence of small amount of data. Also, these densities could be complex and multi-modal, making learning $p(x|y_j)$ from data challenging. Imagine we have a density of the form \cite{SM_Yang}:

\begin{equation} \label{Eq:exponential_density}
    p_\theta(x) = \frac{e^{-f_\theta(x)}}{Z_\theta}
\end{equation}

\noindent where $f_\theta(x)$, parameterized by $\theta$, is an arbitrary function of $x$ (linear or non-linear, e.g. GLMs or NNs), and it's squashed through the exponent to yield positive values, and then normalised by the constant (\textit{w.r.t.} $x$, called \textit{evidence} in a Bayesian posterior density) $Z_\theta = \int e^{-f_\theta(x)} dx$, which gives the regularity requirement for a proper density to integrate to unity. An example is the one-dimensional (1D) Gaussian distribution where $f_\theta(x)=(x-\mu)^2/2\sigma^2, Z_\theta=\sqrt{2\pi}\sigma$ with $\theta=(\mu,\sigma)$.

The density estimation task is to estimate the unknown parameters $\theta$ in $p_\theta(x)$. If distributional structure on data is assumed, e.g. samples are Gaussian distributed (as in LDA and QDA, with further assumptions such as covariance homoscedasticity), we are able to empirically obtain from samples an educated guess of the population summary statistics (e.g. mean and variance) as functions of $\theta$. These assumptions, however, are strong and may lead to poor classifier performance when they are violated. If we have i.i.d samples, we can maximize the data likelihood or minimize some distance metric (e.g. KL divergence) to find $\theta$. However, likelihood-based methods normally require exact computation of the density, which requires knowledge about the normalising constant that may also depend on $\theta$, and in many cases this is intractable. Maximum likelihood estimation (MLE), for example, finds $\theta$ by maximizing the overall likelihood (assuming we have in total $N$ i.i.d samples): 

\begin{equation} \label{Eq:MLE}
    \hat{\theta} = \argmax_{\theta} L(\theta, x) = \argmax_{\theta} \prod_{i=1}^N p_{\theta}(x_i)  
\end{equation}

\noindent or equivalently maximizing the sum of log likelihoods:

\begin{equation} \tag{\ref{Eq:MLE}b}
    \hat{\theta} = \argmax_{\theta} \sum_{i=1}^N \log p_{\theta}(x_i)  
\end{equation}

\noindent where $L(\theta,x)$ is the likelihood function. $\hat{\theta}$ \footnote{A hat (crown) over a variable denotes estimated value.} can be found theoretically by solving the necessary optimal equation $\partial L(\theta, x)/\partial \theta = 0$. As the normalising constant $Z_\theta$ may also be a function of $\theta$, problem arises when exact likelihood computation is not feasible, i.e. we can only evaluate the density from data up to a multiplicative constant. Eq.\ref{Eq:exponential_density} as an example, very often we have only knowledge about $f_\theta(x)$ but not $Z_\theta$. In many cases, particularly high dimensions, the normalising constant $Z_\theta$ cannot be analytically obtained, e.g. due to intractable or expensive integration. We could, of course, resort to numerical evaluation methods such as Monte-Carlo (MC), which may further raise questions about efficiency, accuracy and convergence. Estimating the gradient of log-density could bypass this.

In the following text, we first introduce the \textit{score} function as an alternative characterization and learning objective of data, accompanied by the gradient learning method \textit{score matching}. This learned function can then be used as a generator in sampling to synthesize samples to assist other discriminative classifiers (e.g. nearest neighbours \footnote{Nearest neighbour classifiers can be generative if class conditional densities are learned, e.g. via KDE \cite{ML_Murphy}}, tree-based, neural network, \textit{etc}), particularly in imbalanced data scenario. We call this type \textit{score-assisted} \footnote{'score-assisted' is used mostly when score methods are used indirectly, e.g. in data augmentation, while 'score-based' is used when directly involved, e.g. in generative classification.} discriminative classification. Either, the learned score function can be used to construct the generative density $p(x|y_j)$ conditioned on an initial (e.g. empirically estimated) density value, which can be plugged into a generative classifier. We call this type \textit{score-based} generative classification. We focus on binary classification task and test both methods using simulated \footnote{The term 'simulated' is used in the scene of new data generation, while 'synthesized' or 'synthetic' data is employed in the context of augmentation of original samples.} and real-world data.

\section{Score-based generative modelling}
\label{Sec:score_based_generative_modelling}

\paragraph{Score-based representation of data distribution}
Instead of directly estimating the \textit{pdf}, which may pose challenge in evaluating the normalisation constant, we could estimate the gradient of the log density which is defined as the \textit{score function} $s_\theta(x)$ \cite{SM_Yang}:

\begin{equation} \label{Eq:score_function_definition_1}
    s_\theta(x) = \nabla_x  \log p_\theta(x)
\end{equation}

\noindent The score function has same dimension input and output; the resulted score field is a \textit{conservative}, irrotational vector field (e.g. vanishing \textit{curl} in three-dimensional Euclidean space), over which line integral is path independent. If the density $p_\theta(x)$ is in the specific form of Eq.\ref{Eq:exponential_density}, we have:

\begin{equation} \label{Eq:score_function_definition_2} \tag{\ref{Eq:score_function_definition_1}b}
    s_\theta(x) = -\nabla_x  f_\theta(x) - \nabla_x  \log Z_\theta = -\nabla_x  f_\theta(x)
\end{equation}

\begin{figure}[ht] 
    \centering
    \subfloat[\centering \textit{pdf}]{{\includegraphics[width=0.40\columnwidth]{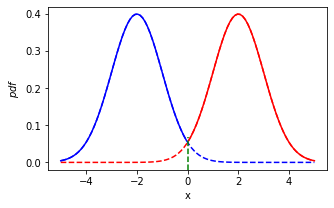}}}
    \qquad
    \subfloat[\centering Score function]{{\includegraphics[width=0.40\columnwidth]{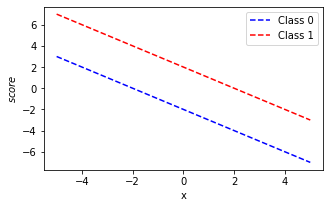}}}
    \caption{Densities and score functions of two 1D Gaussian distributions.}
    \label{Fig:1D_normal_density_and_score_func}
\end{figure} 

\noindent The second term vanishes as the normalising constant $Z_\theta$ is independent of $x$. $s_\theta(x)$ is still a function of $x$, parameterized by $\theta$, with the normalisation constant removed, which is desirable in computation. A score function can be used to characterize a random variable without losing information; it can be parameterized explicitly by any admissible formula, or represented by a composite architecture such as neural network. It admits a scalar-valued or vector-valued $x$ and outputs same-dimension scalar or vector gradient field. 

\begin{figure}[ht] 
    \centering
    \subfloat[\centering \textit{pdf}]{{\includegraphics[width=0.45\columnwidth]{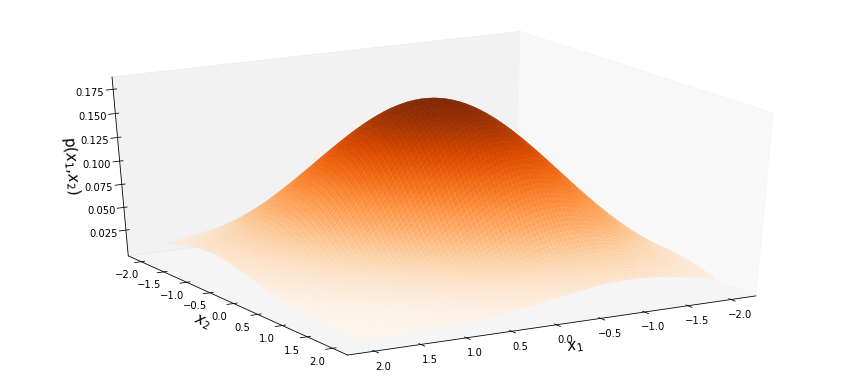}}}
    \subfloat[\centering Scores]{{\includegraphics[width=0.27\columnwidth]{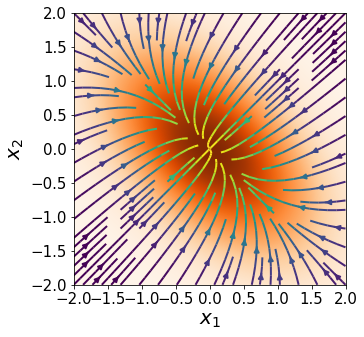}}}
    \caption{Density and score field of a 2D Gaussian distribution.}
    \label{Fig:2D_normal_density_and_score_func}
\end{figure}

As an example, the score function, $s(x)= (\mu-x)/\sigma^2$, of two Gaussian densities with $\mu_0$=-2,$\sigma_0$=1 and $\mu_1$=2, $\sigma_1$=1, are graphically shown in Fig.\ref{Fig:1D_normal_density_and_score_func}. For a general multivariate Gaussian variable $x \sim \mathcal{N}(\mu, \Sigma)$ with density:

\begin{equation} \label{Eq:MND_pdf}
    p(x) = \frac{1}{(2\pi)^{d/2} | \Sigma |^{1/2}} exp[-\frac{1}{2}(x-\mu)^T \Sigma^{-1} (x-\mu)]
\end{equation}

\noindent its score function $s(x|\mu,\Sigma)=\Sigma^{-1}(\mu-x)$. A two-dimensional example with $s_{\theta}(x): \mathbb{R}^2 \rightarrow \mathbb{R}^2$ and $\mu=(0,0), \Sigma=
\begin{bmatrix}
1 & -0.5 \\
-0.5 & 1
\end{bmatrix}
$ is shown in Fig.\ref{Fig:2D_normal_density_and_score_func}.

\begin{figure}[ht]
    \centering
    \includegraphics[width=0.32\linewidth]{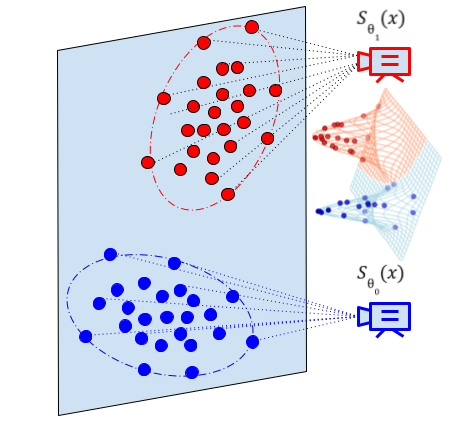}
    \caption{Samples and score generators.}
    \label{fig:score_function_diagram}
\end{figure}

The task of generative modelling is to learn a generative model, e.g. $p_\theta(x)$ or $s_\theta(x)$, from data and to generate new samples using the learned model (Fig.\ref{fig:score_function_diagram}). The key lies in learning an approximate, sample-based $s_\theta(x)$. Suppose we have i.i.d samples $\{x_i \in \mathbb{R}^d \}_{i=1}^N$ drawn from an unknown distribution $p_{D}(x)$ supported by $\chi$, and we want to learn a score function $s_\theta(x): \mathbb{R}^d \rightarrow \mathbb{R}^d$ from data which approximates $p_D(x)$ \cite{SM_Yang}. If we know the density $p_D(x)$ of the data generating process (DGP), we can of course calculate the population-based scores $\nabla_x \log p_D(x_i), i=1,2,...N$, and estimate $\theta$ by minimizing a certain distributional distance (e.g. \textit{KL Divergence}) between $\nabla_x \log p_D(x)$ and $s_\theta(x)$. By doing this we are essentially performing functional approximation. However, we don't have access to $p_D(x)$; in fact, that's the density we are interested in estimating or approximating. \textit{Score matching} \cite{scoreMatching_Hyvarinen1} solves this: it doesn't require access to $p_D(x)$ but optimizes $\theta$ purely based on data. 

\paragraph{Score matching (SM)} 

Hyv\"arinen \cite{scoreMatching_Hyvarinen1} proposed that, non-normalized statistical models can be estimated by minimizing the expected squared distance between the gradient of the log-density given by the model and that of data. Formally, the objective is to find the parameters $\theta$ that minimize the \textit{implicit Fisher Divergence} between the unknown data density $p_{D}$ and approximate density $p_\theta(x)$ \cite{SM_Yang, scoreMatching_Lorenzo}:

\begin{equation} \label{Eq:score_matching_1}
    \theta = \argmin_{\theta} D_F(p_{D} \lVert p_{\theta}) 
\end{equation}

\noindent with

\begin{equation} \label{Eq:score_matching_2}
\begin{aligned} 
    D_F(p_{D} \lVert p_{\theta}) 
    &= \frac{1}{2} \int_\chi p_D(x) \lVert \nabla_x \log p_D(x) - \nabla_x \log p_{\theta}(x) \lVert_2^2 dx \\
    &= \frac{1}{2}\mathbb{E}_{p_D} [\lVert \nabla_x \log p_D(x) - s_{\theta}(x) \lVert_2^2]
\end{aligned}
\end{equation}

\noindent The normalising constants in $p_D$ and $p_{\theta}$ are eliminated when taking derivative, it's thus irrelevant during score matching. Under mild conditions such as differentiability \cite{scoreMatching_Lorenzo},  the \textit{implicit Fisher Divergence} objective can be transformed into its \textit{explicit} form which can be conveniently estimated via sampling routines \footnote{An example implementation can be found at e.g. \cite{SM_Bordyuh}} \cite{scoreMatching_Lorenzo, SM_Yang}:

\begin{equation} \label{Eq:score_matching_2b} \tag{\ref{Eq:score_matching_2}b}
\begin{aligned} 
    D_F(p_D\lVert p_{\theta}) 
    &= \int_\chi p_D(x) \sum_{i=1}^d [\frac{1}{2} (\frac{\partial \log p_\theta (x)}{\partial x_i})^2 + (\frac{\partial^2 \log p_\theta (x)}{\partial x_i^2})^2] dx + C \\
    &= \mathbb{E}_{p_D} [\frac{1}{2} \lVert s_\theta(x) \lVert_2^2 + tr(\nabla_x s_\theta(x))] + C
\end{aligned}
\end{equation}

\noindent where $d$ is the dimension of $x$, $C$ is a constant\footnote{Capital $C$ is used in multiple scenes across this work, e.g. later in integrating the score function to recover density. Its meaning should be clear from context.}. It is proved that \cite{scoreMatching_Lorenzo}, if $p_\theta(x)>0$  for all $x \in \chi$, $D_F(p_\theta \lVert p_{\theta}) \Leftrightarrow p_{\theta}=p_\theta$. 

When computing the \textit{explicit Fisher Divergence} in SM, substantial cost is induced by the Hessian term. Sliced Score Matching (SSM), which measures the divergence over random projections instead, is proposed \cite{SSM_Yang,scoreMatching_Lorenzo}: 

\begin{equation} \label{Eq:score_matching_3}
    D_{FS}(p_D \lVert p_{\theta}) = \frac{1}{2} \int_{\mathcal{V}} q(v) \int_\chi p_D(x) [v^T\nabla_x \log p_D(x) - v^T\nabla_x \log p_{\theta}(x)]^2 dx dv
\end{equation}

\noindent where $v \in \mathcal{V} \subseteq \mathbb{R}^d$ is a noise vector distributed as $q(v)$. Similar to SM, Eq.\ref{Eq:score_matching_3} can be further explicitly developed as \cite{scoreMatching_Lorenzo}:

\begin{equation}  \label{Eq:score_matching_3b} \tag{\ref{Eq:score_matching_3}b}
\begin{aligned} 
    D_{FS}(p_D \lVert p_{\theta}) 
    &= \int_\chi p_D(x) \{\int_{\mathcal{V}} q(v) [v^T(H_x \log p_{\theta}(x))v] dv + \frac{1}{2} \lVert \nabla_x \log p_{\theta}(x) \lVert_2^2 \} dx + C \\
    &= \mathbb{E}_{p_D} \mathbb{E}_{q} [v^T\nabla_x s_\theta(x) v + \frac{1}{2} \lVert s_\theta(x) \lVert_2^2] + C
\end{aligned}
\end{equation}

\noindent where $H_x$ is the Hessian matrix \textit{w.r.t.} coordinate $x$.

Gradient learning approaches such as SM or SSM don't require evaluating the normalizing constant; we can use any parameterized formula or architecture to represent $s_\theta(x)$. In this work, we employ a neural network regressor as our score model and train it using Eq.\ref{Eq:score_matching_2b}.

\paragraph{Constructing density from scores}

Theoretically, once a score function which provides gradient information about the log density everywhere has been learned, we can recover the density:

\begin{equation} \label{pdf_reconstruction_from_score_func}
    p_\theta(x) = C e^{\int_\chi s_{\theta}(x) dx}
\end{equation}

\noindent where $\chi$ is the set supporting $x$, $C$ can determined by any initial condition. Depending on the dimension of $x$, the integral could be over a scalar field or a vector filed (e.g. line integral). With examples of Gaussian densities, we start with one-dimensional $x$, recovering the scalar-valued $p(x)$ from scalar-valued $s_\theta(x)$; then move to two dimensions to recover the scalar field $p(x)$ from a gradient vector field $s_\theta(x)$ via line integration. For writing convenience, we drop the subscript parameter $\theta$ wherever it's clear from the context (e.g. when derivations are parallel for both classes), assuming all functions are parameterized by default (e.g. weights of neural network when representing a score function).

Empirically if we know one point density $p(x_0)$, starting from it we can incrementally construct the density curve (or surface) by visiting many grid points. For example,

\begin{equation} \label{pdf_reconstruction_from_score_func_1D}
    \log p(x_0+\delta x) - \log p(x_0) = \log \frac{p(x_0+\delta x)}{p(x_0)} = \int_{x_0}^{x_0+\delta x} s(x) dx 
\end{equation}

\noindent For one-dimensional $x$, $s(x)$ is also scalar-valued. The integral can be conveniently estimated by numerical integration routines such as Monte Carlo, i.e. $\int_{x_0}^{x_0+\delta x} s(x) dx \approx \frac{\delta x}{N} \sum_{k=1}^{N} s(x_k)$, where $x_k$ are sampled from the interval [$x_0, x_0 + \delta x$]. Essentially, the density at $x_0+\delta x$ can be approximated as: 

\begin{equation} \label{Eq:reconstruct_pdf_from_scores}
    p(x_0+\delta x) = p(x_0) \times e^{\frac{\delta x}{N} \sum_{k=1}^{N} s(x_k)}, 
    \text{where } x_k \in [x_0, x_0 + \delta x]
\end{equation}

\noindent Alternatively, we can use first-order Taylor expansion to approximate the difference, if the step $\delta x$ is small: 

\begin{equation} \label{Eq:reconstruct_pdf_from_scores_Taylor}
    \log p(x_0+\delta x) - \log p(x_0) \approx \frac{d \log p(x)}{dx} |_{x=x_0} \delta x = s(x_0) \delta x
\end{equation}

\noindent The difference between Eq.\ref{Eq:reconstruct_pdf_from_scores} and Eq.\ref{Eq:reconstruct_pdf_from_scores_Taylor} is that, Monte-Carlo method samples a number of points  within the interval, either in a random manner or taking into account the shape of $s(x)$, and average their scores, while Taylor approximation uses one point. There are other integration approximation methods, e.g. trapezoidal rule, Simpson's rule or Gaussian quadrature, that can lead to more accurate computation. Here we take the MC view.

\begin{figure}[ht] 
    \centering
    \subfloat[\centering Class 0 (JSD: 0.01)]{{\includegraphics[width=0.35\columnwidth]{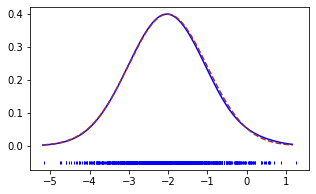}}}
    \qquad
    \subfloat[\centering Class 1 (JSD: 0.07)]{{\includegraphics[width=0.35\columnwidth]{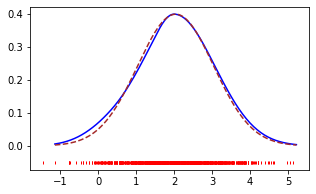}}}
    \\
    \includegraphics[width=0.34\columnwidth]{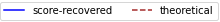}
    \caption{Recovered densities using learned score function. There are 1000 samples for each class, with locations shown at bottom. The score functions use an MLP with size [1, 128, 256, 128, 1] and ReLU activations. Initial probability $p_{\theta_0}(\hat{\mu}_0=[-2.01])=0.40,p_{\theta_1}(\hat{\mu}_1=[2.04])=0.40$.}
    \label{fig:1D_Gaussian_pdfs_recovered_from_score_func}
\end{figure}

As an example, the densities in Fig.\ref{Fig:1D_normal_density_and_score_func} are reconstructed in Fig.\ref{fig:1D_Gaussian_pdfs_recovered_from_score_func} from the learned score fields. The density of Class 0 is better recovered with a smaller Jensen–Shannon divergence (JSD, a symmetric measure of dis-similarity between two distributions) value than Class 1; the difference could have been induced by randomness in samples distribution (e.g. different patterns of sample scarcity at far ends). While the score learning is not perfect, we could improve the learning if we make additional assumptions, e.g. assuming the two clusters of observations are Gaussian distributed, this limits our selection of score architecture to be within the space of linear functions (since we know the score function of a 1D Gaussian density is $s(x)=(\mu-x)/\sigma$, however, there is no reason to prevent ourselves from using more complex functions which might overfit in score matching). This choice of simple score function enables deriving potential analytical closed form of $f(x)=\int_\chi s(x)dx+C$ where $C$ can be determined by some initial condition (e.g. initial probability at a known point, free to choose the cluster center where $p(\hat{\mu})=1/\sqrt{2\pi} \hat{\sigma}$), which could further leads to expressive form of a estimated density. Imposing this extra assumption is discussed in Section.\ref{Sec:score-based generative classification} and results deployed in Fig.\ref{fig:1D_Gaussian_pdfs_recovered_from_score_func_assumeGaussian}.

Same principles can be applied in high dimensions to recover the vector-input, scalar-output \textit{pdf}, from a vector-valued input and output score function $s(x)$. The difference lies in the way we do the integration in Eq.\ref{pdf_reconstruction_from_score_func}: for a vector field, the integration in the exponent is interpreted as line integral:

\begin{equation} \label{pdf_reconstruction_from_score_func_2D}
    \log \frac{p(x_0+\delta x)}{p(x_0)} = \int_{l} s(x) dx = \int_{t_0}^{t_0+\delta t} s[x(t)] \cdot x'(t) dt
\end{equation}

\noindent where $|\cdot|$ denotes inner product. $l: t_0 \rightarrow t_0+\delta t$ is a directed, piecewise smooth curve parameterized by $t$, starting at $x_0$ and ending $x_0+\delta x$ with correspondences $x(t_0)=x_0$ and $x(t_0+\delta t)=x_0+\delta x$. Line integrals over the score vector fields reply on the direction of $x(t)$ but are independent of the parametrization $x(t)$ in absolute value. By definition, the line integral in Eq.\ref{pdf_reconstruction_from_score_func_2D} can be calculated as a Riemann sum of the dot product of the tangent $x'(x)$ and the vector field direction $s[x(t)]$, with infinitesimal partitions over $[t_0,t_0+\delta t]$:

\begin{equation} \label{pdf_reconstruction_from_score_func_2D_integral_0}
    \int_{t_0}^{t_0+\delta t} s[x(t)] x'(t) dt = \lim_{\Delta t \rightarrow 0} \sum_{k=1}^{N} s[x(t_k)] \cdot x'(t_k) \Delta t
\end{equation}

\noindent with $[t_1 = t_0, t_N = t_0 + \delta t]$ and increment $\Delta t$. As per its definition Eq.\ref{Eq:score_function_definition_1}, the score function $s(x)$ is the gradient of a scalar field $p(x)$, i.e. $s(x)$ is a \textit{conservative} vector field, its line integral is path independent, i.e. while there are infinitely many paths between $[t_0, t_0+\delta t]$, we can choose the simplest path, i.e. the straight line connecting the starting and end points, then the tangent $x'(t_k)=[x(t_k)-x(t_{k-1})]/\Delta t$. Further, when $\Delta t$ is sufficiently small, constant score field (e.g. using central point score) can be assumed within $\Delta t$. Together, we have following simplification for contrasting two densities: 

\begin{equation} \label{pdf_reconstruction_from_score_func_2D_integral}
    \log \frac{p(x_0+\delta x)}{p(x_0)}
    = \lim_{\Delta t \rightarrow 0} \sum_{k=2}^{N} \frac{s[x(t_k)]+s[x(t_{k-1})]}{2} \cdot [x(t_k)-x(t_{k-1})]
\end{equation}

\begin{figure}[t] 
    \centering
    \includegraphics[width=0.32\linewidth]{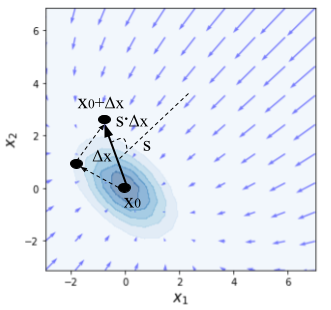}
    \caption{Illustration of the path independent property of line integral. Light blue colored area implies high density arena of Class 0; blue arrows suggests score field. $x_0$ is the starting point, $\Delta x$ the step size. The two paths indicated by black solid arrow and dashed arrows yield the same integration value due to the path independent property.} 
    \label{fig:2D_line_integral}
\end{figure}

An exaggerated illustration of Eq.\ref{pdf_reconstruction_from_score_func_2D_integral}, with one large step contribution, is shown in Fig.\ref{fig:2D_line_integral}. In practice, we start from an initial point $[x_0,p(x_0)]$ with $p(x_0)$ estimated from data (e.g. sampling from the score function using \textit{Langevin dynamics}\footnote{Introduced later in Section.\ref{Sec:examples_discriminative_classification}.}, count the frequencies in the neighbourhood of $x_0$ and compute an estimated density), choose a next point $x_0+\Delta x$, measure the alignment between local score field direction and the trajectory direction, and assign the new point a probability proportional to the exponential cosine similarity. We repeat this incrementally until reaching the target point $x_0+\delta x$, at which point we calculate the final probability as the accumulated contribution from each segment of the trajectory. The whole density surface can be constructed by wandering through the feature space. Note that, the density is constructed with only knowledge of the score function and an initial probability guess; it doesn't require any distributional assumption of data. 

\begin{figure}[ht] 
    \centering
    \subfloat[\centering DGP densities and sample distributions]{{\includegraphics[width=0.45\columnwidth]{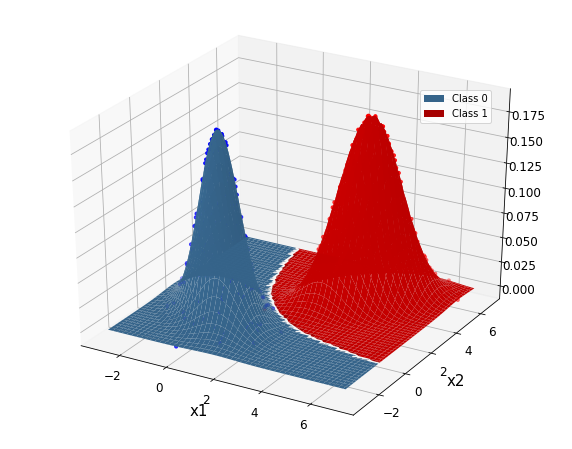}}}
    \qquad
    \subfloat[\centering 2D projections]{{\includegraphics[width=0.33\columnwidth]{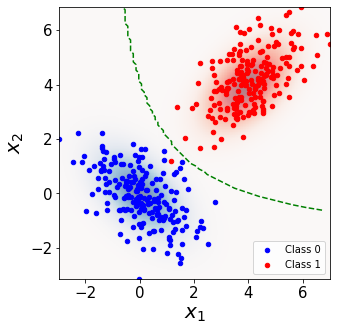}}}
    \caption{2D Gaussian DGPs and simulated samples. In both figures, dots represent samples; on the right, green line marks the boundary with equal density.}
    \label{fig:2D_Gaussian_DGP}
\end{figure}

\begin{figure}[ht] 
    \centering
    \subfloat[\centering Score-recovered densities]{{\includegraphics[width=0.43\columnwidth]{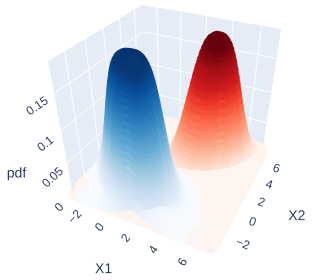}}}
    \qquad
    \subfloat[\centering 2D projections]{{\includegraphics[width=0.35\columnwidth]{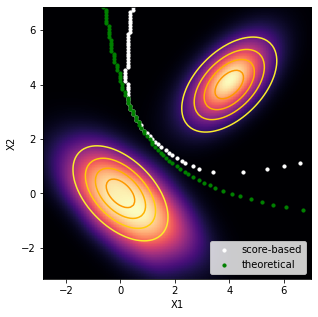}}}
    \caption{Recovered densities and decision boundary using learned score function. There are 200 samples for each class. The score functions use a MLP with size [2, 128, 256, 128, 2] and ReLU activation. Initial probability $p_{\theta_0}(\hat{\mu}_0=[0.07,-0.12])=0.18, p_{\theta_1}(\hat{\mu}_1=[4.06,4.18])=0.18$. On the right, brighter color implies higher recovered density, contour and green dots represent DGP-based (theoretical), white dots indicate the equal density boundary generated by recovered densities.}
    \label{fig:Gaussian_pdfs_recovered_by_scores}
\end{figure}

As an example, we simulated two Gaussian 2D clusters (Fig.\ref{fig:2D_Gaussian_DGP}), each with 200 samples labeled 0/1, with the underlying \textit{DGP}s: $\mu_0=(0,0), \Sigma_0=
\begin{bmatrix}
1 & -0.5 \\
-0.5 & 1
\end{bmatrix}
$,
$\mu_1=(4,4), \Sigma_1=
\begin{bmatrix}
1 & 0.5 \\
0.5 & 1
\end{bmatrix}
$. Two score functions are trained on the simulated samples separately; they are then used to construct the density surfaces in Fig.\ref{fig:Gaussian_pdfs_recovered_by_scores}(a), using estimated probabilities $p_{\theta_0}(\hat{\mu}_0=[0.07,-0.12])=0.18$ and $p_{\theta_1}(\hat{\mu}_1=[4.06,4.18])=0.18$. These estimates of initial probabilities can be obtained either by sampling from the learned score function and counting class appearances in a specified $\sigma$-neighbourhood, or using $p(\hat{\mu})=1/(2\pi)^{d/2}|\hat{\Sigma}|^{1/2}$ if extra Gaussian distribution of data is assumed. It is seen that, the score-recovered densities are smooth and similar to the original ones, except that blue class density is fatter than its DGP density, which might be induced by inaccurate estimations of initial values and score functions. Shown in Fig.\ref{fig:Gaussian_pdfs_recovered_by_scores}(b) is the projected 2D densities and equal density boundaries (which is used later in decision theory). It is observed that, the score-recovered densities co-locate well with theoretical contours; the empirical boundary (white dots) doesn't align fully with the theoretical boundary (green dots) at far ends, this may be a direct result of the excessive power brought by the heavy tail exerted by the over-fat blue class density, or numerical errors at low density areas, as well as incapable extrapolation of the learned score functions in sparse data regions. An improved score-recovered density, with additional assumption of Gaussian distribution of data, is presented in Fig.\ref{fig:Gaussian_pdfs_recovered_by_scores_assumeGaussian}.  

We could generalize the 1D and 2D (and high-dimensional) cases. Eq.\ref{Eq:reconstruct_pdf_from_scores} and Eq.\ref{pdf_reconstruction_from_score_func_2D_integral} are intrinsically the same, i.e. if we divide the interval $\delta x$ into $N$ equidistant intervals $x(t_k)-x(t_{k-1})$ and evaluate $s(x)$ it the middle point. They differ in how we process the product $\delta x \times s(x)$ in the exponent: for one-dimensional $x$ it's scalar product, for $N$-dimensional problem it becomes inner product. We summarize the density construction procedure in Algorithm.\ref{algo:reconstruct_pdf_from_score_func}.

\begin{algorithm} 
\caption{Score-based density construction}
\label{algo:reconstruct_pdf_from_score_func}
\textit{Initialisation}. \linebreak
    $\bullet$ obtain an initial probability at arbitrary point $x_0$. \linebreak
    e.g. $p(x_0=0)=\frac{1}{\sqrt{2\pi}\hat{\sigma}}e^{-\frac{\hat{\mu}^2}{2\hat{\sigma}^2}}$ for 1D Gaussian data, with $\hat{\mu}$ and $\hat{\sigma}$ estimated from data; \linebreak
    \\
\textit{Learning a score function $s(x)$ via score matching}. \linebreak
    \\
\textit{Start from an initial point $[x_0, p(x_0)]$, visit all desired points $[x_0+\delta x, p(x_0+\delta x)]$ by varying $\delta x$. To calculate $p(x_0+\delta x)$:} \linebreak
    $\bullet$ Monte-Carlo sample $N$ points (or with fixed step size) from $[x_0, x_0+\delta x]$. \linebreak
    $\bullet$ compute $p(x_0 + \delta x)$ using Eq.\ref{Eq:reconstruct_pdf_from_scores}.
    \\
\end{algorithm}

\paragraph{Score smoothing} Depending on the parametric form or architecture used to represent a score function, sometimes the learned score surface can be discontinuous. In such cases, we can use for example \textit{kernel smoothing} technique to smooth the estimated function in a post-estimation stage. A simple strategy is to use an identity kernel, i.e. taking simple average of neighboring scores. This post-estimation step equivalently adds locally smooth constraint to the score function, which may be useful in sparse data region where limited information can be learned.

Given the coordinates of an arbitrary point $x^+$, we query the learned score functions to output scores at surrounding points and compute their average (or weighted average) as the central point score. This has been used in constructing the \textit{pdf}, where we sample the integration interval $\delta t$ and take average in Eq.\ref{Eq:reconstruct_pdf_from_scores}. This can further extended to high dimensions where samples are taken from a neighbourhood space:

\begin{equation} \label{Eq:pdf_ratio_func_MC}
    \int_{x_0}^{x_0+\delta x} s(x) dx 
    \approx \frac{V}{N} \sum_{i=1}^{N} s_{\theta_1}(x_i), \text{ where } x_i \in \{x_i: dis(x_i, x^+) \leq \sigma \}
\end{equation}

\noindent where $x_i$ are drawn from the $\sigma$-neighbourhood of $x^+$, $V$ is the volume of the $\sigma$-neighbourhood, $dis(\cdot,x^+)$ is a distance metric associated with the space. Geometry-aware sampling methods (e.g. importance sampling) can be used to sample neighborhood points. Eq.\ref{Eq:pdf_ratio_func_MC} inspires the use of a patch camera centered at $x^+$ with radius (or side length) $\sigma$: we take $N$ points surrounding $x^+$, substitute their scores into Eq.\ref{Eq:pdf_ratio_func_MC}; once $\sigma$ becomes in infinitesimal while $N$ becomes large, the approximation would converge as per law of large numbers.

\section{Score-assisted discriminative classification} \label{Sec:score_assisted_discriminative_classification}

We discuss two applications of gradient learning in discriminative classification: first, in the post-classification stage, we could gain some insights on classification results by examining the score fields of the discriminative density $p(y|x)$ learned by a classifier (e.g. logistic regression); second, we investigate how score-generated samples can be used to enhance other classifiers' performance.

\subsection{Score field of learned discriminative densities} 

\paragraph{Representation of discriminative densities}

In discriminative classification, we directly model the discriminative densities $p(y_j|x)=\frac{e^{-f_{\theta_j}(x)}}{Z_{\theta}}$, where $\theta=\{\theta_1,\theta_2,...,\theta_c\}$, $c$ is the total number of classes. In general, for multi-class classification problem, we use \textit{softmax} probability where $Z_{\theta}=\sum_{j=1}^c e^{-f_{\theta_j}(x)}$, and $f_{\theta_j}(x)$ is typically linear in $\theta_j$; for binary problem, e.g. logistic regression with labels $y \in \{0,1\}$, $f_{\theta_0}(x) = \theta^Tx$, $f_{\theta_1}(x) = 0$ and $Z_{\theta}=1+e^{-\theta^Tx}$. Without loss of generality, we write the binary discriminative densities as $p(y=0|x)=\frac{e^{-f_{\theta_0}(x)}}{Z_{\theta}}$ and $p(y=1|x)=\frac{e^{-f_{\theta_1}(x)}}{Z_{\theta}}$, where $Z_\theta(x)=e^{-f_{\theta_0}(x)} + e^{-f_{\theta_1}(x)}$ (here we have made $Z_\theta$ a function of $\theta$ and $x$), then the score function can be derived as (derivations see Appendix.\ref{App:score_of_discriminative_densities}): 

\begin{equation} \label{Eq:binary_density_score_func}
    s_{\theta_0}(x) = \frac{[f'_{\theta_1}(x) - f'_{\theta_0}(x)]e^{-f_{\theta_1}(x)}}{e^{-f_{\theta_0}(x)} + e^{-f_{\theta_1}(x)}},
    s_{\theta_1}(x) = \frac{[f'_{\theta_0}(x) - f'_{\theta_1}(x)]e^{-f_{\theta_0}(x)}}{e^{-f_{\theta_0}(x)} + e^{-f_{\theta_1}(x)}}
\end{equation}
\noindent where $f'(x)$ denotes derivative \textit{w.r.t.} $x$. Logistic density, for example, yields:

\begin{equation} \label{Eq:logistic_density_score_func}
    s_{\theta_0} = -\frac{\theta'}{1+e^{-\theta^Tx}}, 
    s_{\theta_1} = \frac{\theta' e^{-\theta^Tx}}{1+e^{-\theta^Tx}}
\end{equation}

\noindent where $\theta'$ equals $\theta$ but with intercept removed after differentiation. 

As a comparison to score function, the gradient of the densities are (Appendix.\ref{App:score_of_discriminative_densities}):

\begin{equation} \label{Eq:binary_density_func}
    \nabla_x p_{\theta_0}(x) = \frac{\nabla_x f_{\theta_1}(x)-\nabla_x f_{\theta_0}(x)}{[1+e^{f_{\theta_0}(x)-f_{\theta_1}(x)}][1+e^{f_{\theta_1}(x)-f_{\theta_0}(x)}]}
\end{equation}
\noindent and $\nabla_x p_{\theta_1}(x) = - \nabla_x p_{\theta_0}(x)$, as $p(y=1|x)=1-p(y=0|x)$ is satisfied everywhere. Logistic density, for example, gives:

\begin{equation} \label{Eq:logistic_density_derivative_func}
    \nabla_x p_{\theta_0}(x) = - \frac{\theta}{(1+e^{\theta^Tx})(1+e^{-\theta^Tx})}
\end{equation}

\paragraph{Decision theory for discriminative classification}

Let $L(y,\hat{y})$ be the loss induced by classify $y$ (ground truth) as $\hat{y}$ (prediction), we have $L(y,\hat{y})=0$ if $\hat{y}=y$, and $L$ can be symmetric or asymmetric to reflect, e.g. in medical cancer scanning or credit card fraud detection scenarios, the impact or operational cost induced by \textit{false positives} (FPs) and \textit{false negatives} (FNs) could be different. A common choice of $L$ is the zero-one loss which simply counts the misclassification (FPs and FNs) numbers. The expected risk of predicting label $\hat{y}$ given $x$ is therefore $R(\hat{y}|x)=\sum_c L(y,\hat{y}) p(y|x)$, and we make the optimal prediction $y^*=\argmin_{\hat{y}} R(\hat{y}|x)$. This optimal decision rule essentially leads to $y^*=\argmax_{y} p(y|x)$, i.e. choosing the class with highest label probability at $x$, as this minimizes the expected loss at $x$ \cite{GP_Rasmussen}, this optimal classifier is known as the \textit{Bayes classifier}. Using this construction, we divide the feature space into multiple decision regions, and we are interested in finding the \textit{decision boundary} where the two probability curves or surfaces meet with equal probability, and we may expect uncertainty to increase near the boundary as the class probabilities approach each other. More detailed treatment of decisions can be found in e.g. \cite{decision_theory_Berger, elements_Friedman, Intro_James, ML_Murphy}. 
\\
In binary classification, we typically use the following decision rule to assign a label $\hat{y}$ to an observation $x$:

\begin{equation} \label{Eq:binary_decision_theory_1}
\hat{y} = 
\begin{cases}
    1, & \text{if } p(y=1|x) - p(y=0|x) \geq \gamma_0 \\
    0,              & \text{otherwise}
\end{cases}
\end{equation}

\noindent where $\gamma_0 \geq 0$ is a \textit{soft margin} which represents a minimum density gap for the two classes to separate. The intuition behind is, given the feature $x$ and the estimated probabilistic model $p(y|x)$, if the probability of being one class is higher than being the other to certain level $\gamma_0$, we can safely classify the sample as the higher probability class. $\gamma_0$ can be set to zero if we are not hoping for high confidence of separation. Rather than measuring absolute difference, we can also use the ratio $p(y=1|x) / p(y=0|x) \geq \gamma_0$ as a discriminative criterion:

\begin{equation} \label{Eq:binary_decision_theory_2} \tag{\ref{Eq:binary_decision_theory_1}b}
\hat{y} = 
\begin{cases}
    1, & \text{if } \frac{e^{-f_{\theta_1}(x)}}{Z_{\theta}} / \frac{e^{-f_{\theta_0}(x)}}{Z_{\theta}} \geq e^{\gamma_0} \\
    0,              & \text{otherwise}
\end{cases}
\end{equation}

A key step in making prediction using Eq.\ref{Eq:binary_decision_theory_2} is to find the decision boundary equation $\frac{e^{-f_{\theta_1}(x)}}{Z_{\theta}} / \frac{e^{-f_{\theta_0}(x)}}{Z_{\theta}} = e^{\gamma_0}$, which represents a curve or surface. Let's define a distance function $\gamma(x)$:

\begin{equation} \label{Eq:d_x}
    \gamma(x) = \log \frac{p(y_1|x)}{p(y_0|x)} - \log e^{\gamma_0} = f_{\theta_0}(x) - f_{\theta_1}(x) - \gamma_0 \\ 
\end{equation}

\noindent By introducing $\gamma(x)$\footnote{Not to be confused with the hyper-parameter discard rate, also denoted by $\gamma$, which is used later in Langevin sampling of imbalanced data.}, the discriminator Eq.\ref{Eq:binary_decision_theory_2} turns into:

\begin{equation} \label{Eq:binary_decision_theory_3} \tag{\ref{Eq:binary_decision_theory_1}c}
\hat{y} = 
\begin{cases}
    1, & \text{if } \gamma(x) \geq 0 \\
    0,              & \text{otherwise}
\end{cases}
\end{equation}

\noindent The classification task becomes finding the roots of $\gamma(x)=0$, which can be done using a numerical root-finding scheme such as the \textit{Newton-Raphson} (NR) method:

\begin{equation} \label{Eq:Newton_Raphson}
    x_{k+1} = x_k - \frac{\gamma(x_k)}{\nabla_x \gamma(x_k)}
\end{equation}

\paragraph{Score field of learned logistic densities} The use of score fields to characterize post-classification densities is illustrated using a toy example in which Gaussian data is separated by a logistic regression classifier. 

\begin{figure}[ht] 
    \centering
    \subfloat[\centering DGP densities and classification results (training)]{{\includegraphics[width=0.38\columnwidth]{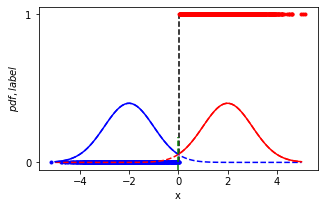}}}
    \qquad
    \subfloat[\centering Zoom-in details of the decision boundary]{{\includegraphics[width=0.36\columnwidth]{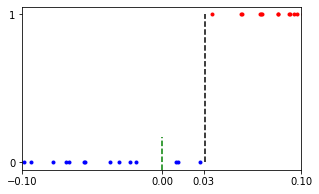}}}
    \\
    \subfloat[\centering Score function of the learned logistic densities]{{\includegraphics[width=0.37\columnwidth]{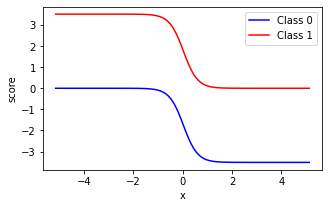}}}
    \qquad
    \subfloat[\centering Gradients of the learned logistic densities]{{\includegraphics[width=0.38\columnwidth]{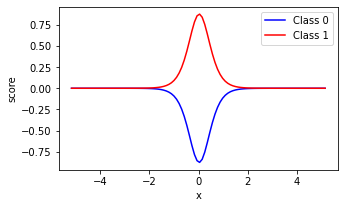}}}
    \caption{Logistic regression classifier applied to toy 1D Gaussian data.}
    \label{fig:1D_Gaussian_logistic_regression}
\end{figure}

We first look at the 1D case. 2000 samples are simulated (1000 for each class) with densities $\mathcal{N}(-2,1)$ and $\mathcal{N}(2,1)$ (same densities as used in Fig.\ref{Fig:1D_normal_density_and_score_func}). Logistic regression (LR) is applied to the data to find the linear discrominator (the \textit{logit} function in Eq.\ref{Eq:sigmoid_func}) $f(x)=f_{\theta_0}(x) = \theta^Tx=\theta_0 + \theta_1x$ via MLE. The data and classification results are shown in Fig.\ref{fig:1D_Gaussian_logistic_regression}, where the learned parameters are $\theta_0$=-0.1, $\theta_1$=3.5. The resulted logistic densities (Eq.\ref{Eq:sigmoid_func}) are $p(y=0|x) = e^{0.1 - 3.5x}/(1+e^{0.1 - 3.5x}), p(y=1|x) = 1/(1+e^{0.1 - 3.5x})$. Using Eq.\ref{Eq:d_x} with $\gamma_0=0$, we obtain the decision boundary $x=-\theta_0/\theta_1=0.03$.

We can obtain the scores of the learned logistic densities by plugging the learned $f_{\theta_0}(x)$ and $f_{\theta_1}(x)$ in to Eq.\ref{Eq:logistic_density_score_func}, as shown in Fig.\ref{fig:1D_Gaussian_logistic_regression}(c). It is observed that, in the region close to the decision boundary $x=0.03$, the two score functions exhibit abrupt changes, and gradients have sharp peaks at the boundary point.

\begin{figure}
    \centering
    \subfloat[\centering LR decision boundary]{{\includegraphics[width=0.31\columnwidth]{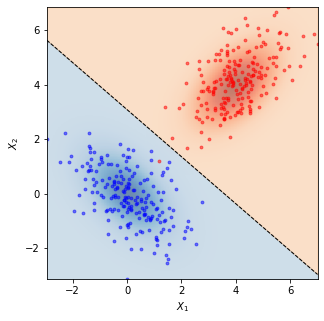}}}
    \subfloat[\centering Scores at sample points]{{\includegraphics[width=0.32\columnwidth]{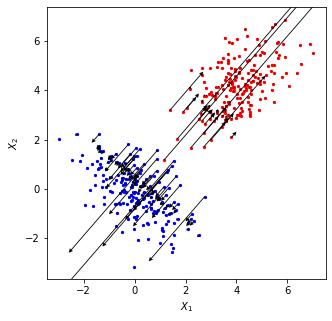}}}
    \subfloat[\centering Gradients at sample points]{{\includegraphics[width=0.32\columnwidth]{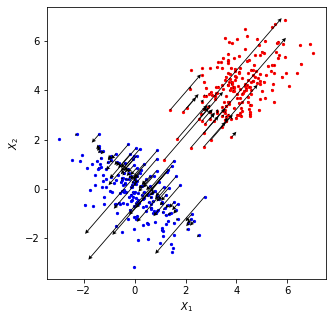}}}
    \caption{A toy logistic regression classifier with 2D Gaussian data.}
    \label{fig:2D_Gaussian_logistic_regression}
\end{figure}

The same procedure is repeated for the simulated 2D Gaussian data from Fig.\ref{fig:2D_Gaussian_DGP}, in which we each have 200 samples for each class. After applying LR, we have the decision boundary equation $\gamma(x)=f(x)=f_{\theta_0}(x) = \theta^Tx=\theta_0+\theta_1x_1+\theta_2x_2=0$ with coefficients $\theta_0=-6.17, \theta_1=1.73, \theta_2=2$ estimated by MLE. The decision boundary is plotted in Fig.\ref{fig:2D_Gaussian_logistic_regression}. Again, we observe large score and gradient values in regions near the boundary line; points far away from the boundary have vanishing scores. The score vector norm may imply some sort of sample weights in making classification decisions (e.g. identifying support vectors).

\subsection{Discriminative classification with score-augmented data}
\label{Sec:examples_discriminative_classification}

A straightforward application of score-based generative modelling is to learn a score function from training data (potentially small and sparse), and use it to synthesize credible samples via gradient-based sampling methods (e.g. \textit{Langevin dynamics}) to to populate the sample space (e.g. augmenting minor class), which can improve the performance of an off-the-shelf discriminative classifiers (e.g. nearest neighbour voting), particularly in imbalanced learning tasks.

\subsubsection{A simulated 2D Gaussian data example}

\begin{figure}[ht]
    \centering
    \includegraphics[width=0.32\columnwidth]{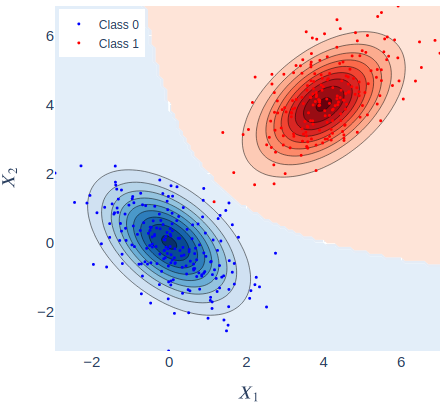}
    \caption{2D Gaussian densities, samples and theoretical boundary}
    \label{fig:2D_Gaussian_classification}
\end{figure}

We use the toy 2D Gaussian data from Fig.\ref{fig:2D_Gaussian_DGP} as an example, and re-draw the two classes, labelled 0 (colored blue) and 1 (colored red), in Fig.\ref{fig:2D_Gaussian_classification}. As the data are relatively well separated, very few cross-class samples exist (i.e. samples interfering with the other class).

\begin{figure}[ht] 
    \centering
    \subfloat[\centering Class 0]{{\includegraphics[width=0.36\columnwidth]{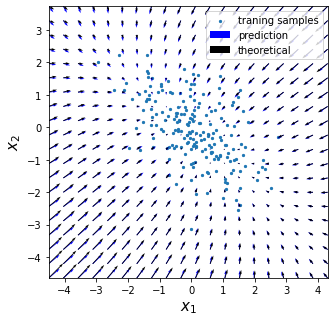}}}
    \qquad
    \subfloat[\centering Class 1]{{\includegraphics[width=0.35\columnwidth]{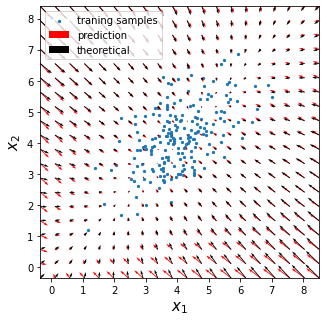}}}
    \caption{Theoretical and predicted scores at grid points.}
    \label{Fig:trained_score_functions}
\end{figure}

\paragraph{Score matching}
We fit two score functions to the samples, one for each class. The score function is represented using an multi-layer perceptron (MLP, fully connected neural network) trained with the loss function in Eq.\ref{Eq:score_matching_2b} \cite{SM_Yang, SM_Bordyuh}. The matching between theoretical and predicted scores is shown in Fig.\ref{Fig:trained_score_functions}, from which we observe better fit in the first class, although marginal deviations between the theoretical and predicted score fields exist in both classes (i.e. gradient flows pointing away at far ends), which implies a tendency for the trained score function to potentially stretch the score field, and biasely draw samples from less populated arenas (which are currently under-represented by the given samples).    

\begin{figure}[!htb]
    \centering
    \begin{minipage}{0.50\columnwidth}
        \centering
        \includegraphics[width=0.70\columnwidth]{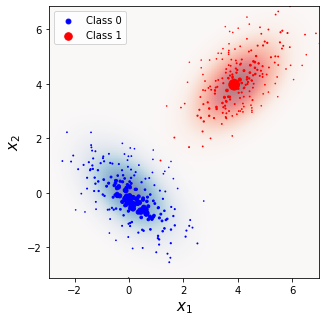}
        \caption{Probabilities (represented by point size) predicted at original samples.}
        \label{fig:score_predicted_at_sample_points}
    \end{minipage}%
    \qquad
    \begin{minipage}{0.43\columnwidth}
        \centering
        \includegraphics[width=0.85\columnwidth]{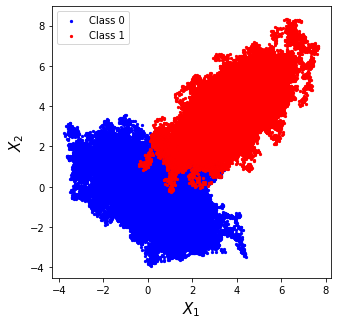}
        \caption{New samples generated using learned score functions.}
        \label{fig:new_sample_generation}
    \end{minipage}
\end{figure}

We observe from the definition of score function (Eq.\ref{Eq:score_function_definition_1}) that, the reciprocal of the norm of a point score is roughly proportional to its \textit{pdf} value, i.e. 

\begin{equation} \label{Eq:score_func_prob}
    | s_\theta(x) | = | \nabla_x  \log p_\theta(x) |
    = |\frac{1}{p_\theta(x)}| \times | \nabla_x p_\theta(x) |
\end{equation}

\noindent which hints we could use the score norm as a (very) rough estimate of probability. We thus ask the trained score functions to predict the scores at training samples and plot the samples with point size proportional to the predicted probability hinted by Eq.\ref{Eq:score_func_prob}, as shown in Fig.\ref{fig:score_predicted_at_sample_points}. The predictions are consistent with our \textit{DGP}s, with central points given higher predicted probabilities. However, caution should be alerted as this rough approximation is by no means accurate, we may miss significant contribution from $| \nabla_x p_\theta(x) |$ in Eq.\ref{Eq:score_func_prob}. 

\paragraph{Sampling with Langevin dynamics} 
The power of generative modelling, either directly modelling the \textit{DGP} (e.g. \textit{GAN}) or learning an underlying density distribution, lies in its capacity of generating plausible, synthesized samples at currently unavailable (e.g. due to data collection costs) regimes by sampling from the learned process (e.g. probabilistic extrapolation). However, sampling from arbitrary density is not easy \cite{Bayesian_signal_processing_Joseph}; numerically it can be approached by Markov chain Monte Carlo (\textit{MCMC}) methods which normally involve a rejection process. As we only have accessible the sample-based score model $s_\theta(x)$, which is an approximator of the gradient of the log-density, we use a specific MCMC procedure called \textit{Langevin dynamics} \cite{Langevin_Welling} to iteratively generate a chain of samples, starting from an initial known sample $x_0$ \cite{SM_Yang}:

\begin{equation} \label{Eq:Langevin_dynamics}
    x_{i+1} = x_i + \frac{\epsilon}{2} \nabla_x\log p_\theta(x_i) + \sqrt{\epsilon} z_i, i=0,1,...,N-1
\end{equation}

\noindent where $z_i \sim \mathcal{N}(\textit{\textbf{0}},\textit{\textbf{I}})$. Following the Langevin dynamics, a random initial sample will move gradually to high density regions following the gradient vector field of log-density. Unlike other \textit{MCMC} methods such as the Metropolis-Hastings sampler \cite{MC_Hastings} which may need access to the (un-normalised) distribution $p_\theta(x)$, sampling using Langevin dynamics only requires the gradient of log-density. If the perturbation parameter $\epsilon$ (also step size) is small and the chain length $N$ is large to achieve equilibrium, the samples generated by Eq.\ref{Eq:Langevin_dynamics} will converge to the true distribution $p_\theta(x)$. Empirically, we can approximate $\nabla_x\log p_\theta(x)$ using $s_\theta(x)$, and repeat the above sampling process for many times, each time randomly starting from an existing sample $x_0$ with probability $p_\theta(x_0) \propto 1/| s_\theta(x) |$ (Eq.\ref{Eq:score_func_prob}). 

Originally, we have 200 samples for each class; we ask the trained score functions to repeatedly generate 100 chains, with 1000 samples per chain ($\epsilon=0.003$) and first 200 samples discarded (ultimately we have in total 80,000 valid samples for each class). The generated samples are shown in Fig.\ref{fig:new_sample_generation}. It is seen that, the density shape is preserved, with more samples clustered in the mean arena and sparse samples at margins. It massively extends the given sample profile, and even populates low density arenas, which is desirable when analysing extreme events or expensive datasets. However, although data sparsity is reduced, we should also be cautious about potentially spurious samples in arenas far from the original clusters. As noted by \cite{SM_Yang}, data scarcity in low density regions can invalidate score estimation and Langevin sampling. This is observed from the training process where deviations between theoretical and predicted scores exist and may lead to a stretched valley (i.e. the learned score function may represent a dynamics different from the true DGP). This effect could be alleviated by using annealed Langevin dynamics \cite{SM_Yang}.

\begin{figure}[ht] 
    \centering
    \subfloat[\centering Contrasting \textit{pdf}s]{{\includegraphics[width=0.33\columnwidth]{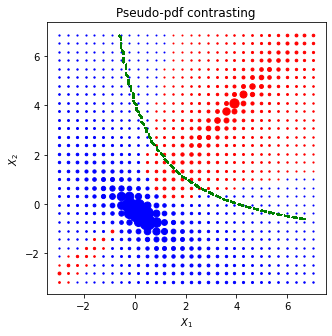}}}
    \subfloat[\centering Fixed radius]{{\includegraphics[width=0.33\columnwidth]{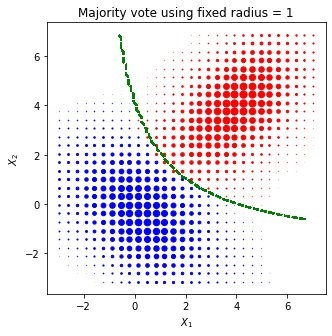}}}
    \subfloat[\centering Fixed nearest neighbour number]{{\includegraphics[width=0.33\columnwidth]{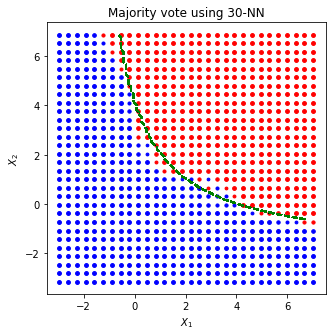}}}
    \caption{Predicted labels at grid points using generated samples (point size represents confidence level, green dashed line denotes theoretical quadratic boundary, Class 0 is colored blue and Class 1 red).} 
    \label{Fig:Gaussian_classification_results_new_samples}
\end{figure}

With the augmented data, we can apply many convenient classification algorithms, hoping for improved performance compared to just using original sparse data. With plausible volume of augmented samples, a naive approach is to label a unknown test sample by majority voting within its $\sigma$-neighbourhood, i.e. using empirical ratio of labels as an estimate for the discriminative probability $p(y|x)$. For locating the neighbourhoods, two strategies can be applied: we can either fix the search radius $\sigma$ and search within a circle centering the test point, or we can search until we find a fixed number of nearest objects (i.e. constant \textit{nearest neighbours}). Another intuitive method for labelling a test point $x=(x_1,x_2)$ is contrasting the approximate \textit{pdf} values (i.e. $p_{\theta_i}(x_0) \propto 1/| s_{\theta_i}(x) |$) of both classes. Results derived using the generated samples are shown in Fig.\ref{Fig:Gaussian_classification_results_new_samples}, where performances of the three classification methods are compared. It is observed that, all three methods are more confident about points close to class centers (confidence is represented by point size, where for \textit{pdf}s, confidence level is proportional to its magnitude, and for counting methods, it's the proportion of points of majority class); points lying at boundary, as evident in the nearest neighbour method, are blurry (indicated by smaller dot size) and thus involve more uncertainties.

The pseudo-\textit{pdf} method labels a data point by contrasting the two inverse score norms, which may be very inaccurate if the ignored contribution of $| \nabla_x p_\theta(x) |$ is significant in Eq.\ref{Eq:score_func_prob}. The two counting-based methods, i.e. fixed radius and fixed nearest neighbours majority voting, are based on the newly generated samples with known labels; the choice of fixed radius and number of nearest neighbours reflects a trade-off between bias and variance (i.e. the classifier's generalisation capacity). It is seen that searching fixed number of nearest neighbours gives the most accurate decision boundary close to the natural boundary given by the DGPs in Fig.\ref{fig:2D_Gaussian_DGP}.

\begin{figure}[ht]
    \centering
    \begin{minipage}{0.35\columnwidth}
    \centering
    {\includegraphics[width=0.95\columnwidth]{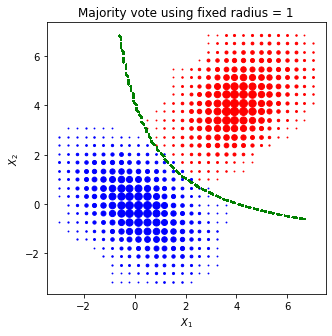}}
    \subcaption{Fixed radius}
    \end{minipage}
    \quad
    \begin{minipage}{0.60\columnwidth}
    \centering
    {\includegraphics[width=0.55\columnwidth]{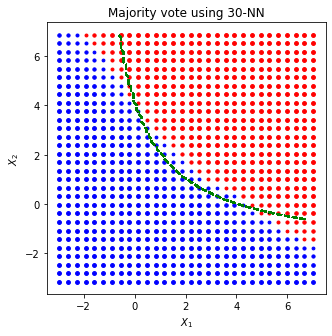}}
    \subcaption{Fixed nearest neighbour number}
    \end{minipage}
    \caption{Predicted labels at grid points using original samples (point size represents confidence level, green dashed line denotes theoretical quadratic boundary, Class 0 is colored blue and Class 1 red).} 
    \label{Fig:Gaussian_classification_results_original_samples}
\end{figure}

\noindent To see the effect of data augmentation, we also present the results built on original small samples in Fig.\ref{Fig:Gaussian_classification_results_original_samples}. We observe narrower and less dense high-confidence arenas in the fixed radius case, and less accurate inference boundary using the constant neighbours method. In both cases, predictions at the far ends (top left and bottom right) are less credible due to lack of training data in the neighbourhoods. Comparing Fig.\ref{Fig:Gaussian_classification_results_new_samples} and Fig.\ref{Fig:Gaussian_classification_results_original_samples} demonstrates the power and effectiveness of score-based data augmentation in enhancing classification quality, in the presence of small, sparse data.

\paragraph{A multi-modal score function}

Instead of learning two score functions separately, we could, however, learn a generic, multi-modal generator which produces samples for both classes. This approach could be more efficient in terms of density estimation, but won't help with classification because a uniform generator can't help populate the feature space with labels, neither can we do \textit{pdf}s contrasting: the score function only gives one unified score for an input $x$, regardless of its class. However, we could potentially utilize the unified score values of the data with known labels as inputs to train a classifier, i.e. classification directly using score features. Note that, this is not a certified approach because there is a risk that, two points from distinct classes may share the same score, e.g. $p_{\theta_1}(x)=Cp_{\theta_0}(x)$, which makes them indistinguishable and misleads the classifier in the learning process.

\begin{figure}[ht]
    \centering
    \begin{minipage}{0.35\columnwidth}
    \centering
    {\includegraphics[width=0.95\columnwidth]{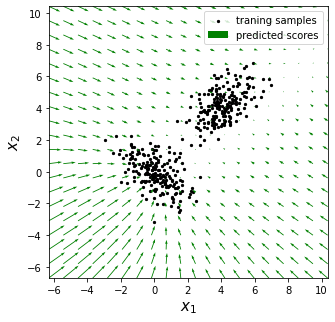}}
    \subcaption{Predicted scores at grid points}
    \end{minipage}
    \quad
    \begin{minipage}{0.60\columnwidth}
    \centering
    {\includegraphics[width=0.55\columnwidth]{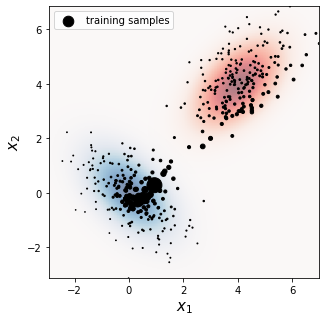}}
    \subcaption{Probabilities (represented by point size) predicted at original samples}
    \end{minipage}
    \caption{Learned multi-modal score function and predicted pseudo probabilities.}
    \label{Fig:Gaussian_classification_generic_score_function}
\end{figure}

As a toy example, the learned multi-modal score function and its capacity to re-produce the training samples are shown in Fig.\ref{Fig:Gaussian_classification_generic_score_function}. The scores predicted by the learned multi-modal score function (Fig.\ref{Fig:Gaussian_classification_generic_score_function}(a)) are reasonable in the sense that, it assigns small score values (i.e. proportionally heavy densities as per Eq.\ref{Eq:score_func_prob}) to the two cluster centers, which is also evidenced from Fig.\ref{Fig:Gaussian_classification_generic_score_function}(b). However, it also puts some undesirable masses on margin points in between the two clusters. This could induce skewed, unrealistic behaviour when using the unified score function to represent the original two clusters with different underlying dynamics. 

\begin{figure}[ht] 
    \centering
    \begin{minipage}{0.35\columnwidth}
    \centering
    {\includegraphics[width=0.95\columnwidth]{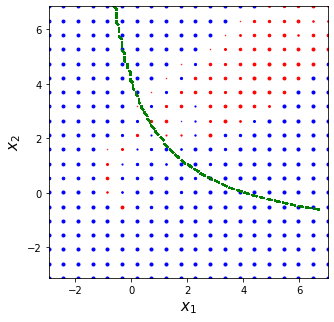}}
    \subcaption{XGB}
    \end{minipage}
    \quad
    \begin{minipage}{0.60\columnwidth}
    \centering
    {\includegraphics[width=0.55\columnwidth]{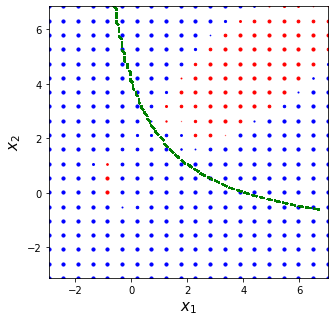}}
    \subcaption{NN}
    \end{minipage}
    \caption{XGB and NN predicted labels at grid points (point size represents confidence level, green dashed line denotes theoretical quadratic boundary, green represents Class 0 and red Class 1). Both classifiers are trained using score features of original samples.} 
    \label{Fig:Gaussian_classification_results_scores}
\end{figure}

As a trial of classifying points based on scores predicted by the unified score funtion, the classification boundaries, sketched using the two popular discriminative classifiers, i.e. \textit{extreme gradient boosting trees} (XGB) and \textit{neural network} (NN), are shown in Fig.\ref{Fig:Gaussian_classification_results_scores}. Both classifiers are trained to map the relation between the generic scores and labels; however, this mapping is biased: most predicted labels favor Class 0 (colored blue), only those in the near cluster center regimes (where density is large large) are credible for Class 1. We thus conclude that scores on its own are insufficient features for label hinting. That said, any method searching nearest neighbours purely based on scores are invalid as well.

\subsubsection{Score-assisted imbalanced learning}

\begin{figure}[ht] 
    \centering
    \includegraphics[width=0.75\columnwidth]{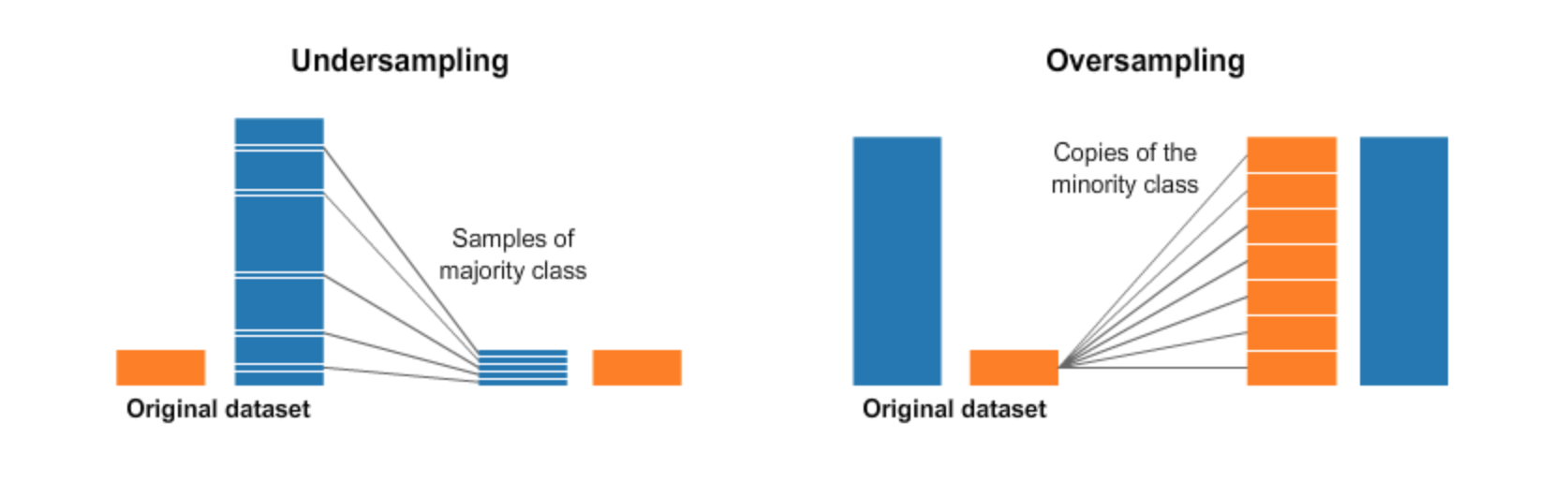}
    \caption{Undersampling and oversampling \cite{credit_card_Will}}
    \label{Fig:undersampling_oversampling}
\end{figure}

In real-world, it's unlikely to have two well separated classes such as the case in Fig.\ref{fig:2D_Gaussian_DGP}; more often we may encounter high dimensional, mixed-classes data (e.g. images), and the amount of available data varies much across classes. This is common in modelling extreme events such as fraud detection (e.g. large claims in insurance, occurrence of fires or floods in natural hazards, \textit{etc}). A classifier trained on imbalanced data are likely to be biased in decision-making: it may tend to acknowledge the class with more exposure. People have been using upsampling (oversampling) and/or downsampling (undersampling) techniques (Fig.\ref{Fig:undersampling_oversampling}) to make a dataset balanced. Here we demonstrate the use of score functions to generate more credible or realistically-like samples for the minor class in a principled manner, hoping that score functions could be better generators in high dimensions, and classifiers trained on score-augmented data could be more skillful in identifying rare cases. This is particularly useful when large volume of data is not available due to, for example, expensive cost in data collection (e.g. clinical trials); instead, we only need to collect a few representative samples over the underlying distribution (i.e. the generative dynamics) and interpolating or extrapolating them as per the learned score function. In this regard, score function may produce more representative synthesized samples by encoding the gradient information of a log-density, and enables more efficient and flexible data generation via Langevin sampling. 

\subsubsection*{A simulated high-dimensional imbalanced data example}

In our first imbalanced example, we simulate in total 3000 ten-dimensional samples with small noise (0.01\% random label flipping), among which 2830 samples are negative (labelled 0) and 170 are positive (labelled 1). The two classes are mixed and highly imbalanced (positive-to-negative ratio $\sim$ 1:95). With more exposure to negative samples, a general classifier trained on this data may tend to label unseen sample as negative. 

\begin{figure}[ht] 
    \centering
    \subfloat[\centering Simulated data (showing first two dimensions)]{{\includegraphics[width=0.40\columnwidth]{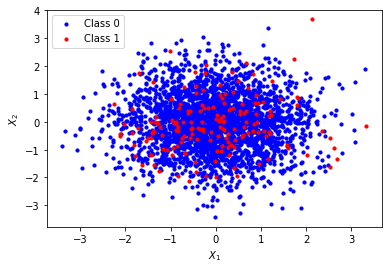}}}
    \qquad
    \subfloat[\centering Class imbalance]{{\includegraphics[width=0.25\columnwidth]{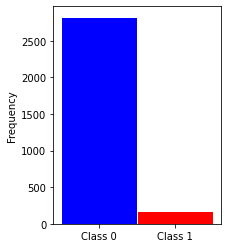}}}
    \caption{Theoretical and predicted scores at grid points.}
    \label{Fig:imbalanced_simulated_data}
\end{figure}

The data is proportionally split (i.e. stratified) into training ($\sim$ 75\%) and test ($\sim$ 25\%) sets, with size of 2259 (positive: 128) and 571 (positive: 42) in each set. While the test set remains fixed, positive samples in the training set are augmented by sampling the learned score function using Langevin dynamics. The score function is trained on the 128 positive training samples via score matching, and then used to generate 2002 positive samples, making numbers of positives and negatives equal after sampling. Then a discriminative classifier is fit to the newly augmented training data, and its performance is reported on the fixed test set.

\begin{table}[ht] 
\small
\centering
\begin{tabular}{p{1.1cm}|p{1.8cm}|p{1.8cm}|p{1.8cm}|p{1.8cm}} 
\multicolumn{2}{c}{}&\multicolumn{2}{c}{Predicted}&\\
\cline{3-4}
\multicolumn{2}{c|}{}&{Negative}&{Positive}&\multicolumn{1}{c}{}\\
\cline{2-4}
\multirow{2}{*}{Actual} &{Negative} &{TN} &{\cellcolor[gray]{0.9}FP ($\alpha$)} & {}\\
\cline{2-4}  
&{Positive} &{\cellcolor[gray]{0.9}FN ($\beta$)} &{\cellcolor[gray]{0.7}TP} & \ding{214} \textit{recall}\\
\cline{2-4}
\multicolumn{1}{c}{} & \multicolumn{1}{c}{} & \multicolumn{1}{c}{\multirow{2}{*}{\parbox{2em}{}}} & \multicolumn{1}{c}{\multirow{2}{*}{\parbox{2em}{\vskip1ex\ding{215}\\ \textit{precision}}}} &\multicolumn{1}{c}{}\\
\end{tabular}
\vspace{0.25cm}
\caption{Illustration of the \textit{confusion matrix} for binary classification.}
\label{Tab:confusion_matrix}
\normalsize
\end{table}

The full examination table, i.e. the \textit{confusion matrix}, is illustrated in Table.\ref{Tab:confusion_matrix}, in which TN denotes true negative, FP false positive, FN false negative, and TP true positive. $\alpha$ refers to type I error, $\beta$ is the type II error. When classifying imbalanced data, rather than focusing on accuracy (which even in worst case could be high, i.e. imagine the classifier just dumbly labels all test samples as negative), we are particularly interested in the sensitivity (true positive rate or recall, a ratio of true positives to all actual positives) and $F_1$ score metrics, the former emphasizes the skill of a classifier to identify all positives, the later is a metric blending precision and recall:

\begin{equation}
    F_1 = \frac{2 \times \textit{\text{precision}} \times \textit{\text{recall}}}{\textit{\text{precision}} + \textit{\text{recall}}}
\end{equation}

\noindent where \textit{precision}=TP/(TP+FP) is the ratio of true positives to number of samples the classifier claims to be positive; \textit{recall}=TP/(TP+FN) is the ratio of true positives to the number of actual positives. They both assess a binary classifier's ability to identify positives, benchmarking on different denominators. 

When generating new samples using the trained score function, apart from the step size parameter $\epsilon$ in Langevin dynamics (Eq.\ref{Eq:Langevin_dynamics}), we add two extra hyper-parameters: chain length $l$ and discard rate $\gamma$ to allow more flexibilities when sampling. Each time we randomly start from one of the 128 existing positive points in the training set, walk $l$ steps and drop the initial $l \times \gamma$ samples in the Langevin chain. Therefore, we can generate new samples with different characteristics using the same score function, e.g. short Langevin walk around existing points and preserving most points along the trajectory, or walking long while keeping only last few points. The newly generated samples are then input into two discriminative classifiers, namely a \textit{random forest} (RF) classifier and a \textit{neural network} (NN) classifier. The results are presented in Table.\ref{Tab:synthesized_imbalanced_discriminative}. Also reported are classification results based on the original imbalanced data, and results from two two popular minority oversampling methods: the synthetic minority over-sampling technique (SMOTE) \cite{SMOTE_Chawla} and the Adaptive Synthetic (ADASYN) method \cite{ADASYN_He}.

It's observed that, classifiers using the score function generated samples consistently outperform those using SMOTE and ADASYN upsampling methods, reporting higher recall and $F_1$ values (and most precisions). An evident example is the score-based random forest classifier with $l=10, \gamma=0.2, \epsilon=0.01$ (colored red in the middle) which results in similar numbers of TNs and FPs as SMOTE and ADASYN, but with significantly smaller number of FNs and larger number of TPs. In the other two score-based settings (the last two scenarios in Table.\ref{Tab:synthesized_imbalanced_discriminative}), we observe record high recall and $F_1$ values, respectively. Also notable is the inferior performance of neural network classifiers as compared to the emsembling method in terms of recall (except the first scenario), this might be due to the complexity of the data and the design of neural network architecture. For example, we use a fully connected MLP with layer sizes [10, 32, 64, 128, 64, 32, 1], ReLU activation functions (except output layer) and binary cross entropy loss objective, trained to maturity with risk of overfitting. 

\vspace{0.5cm}
\begin{minipage}{0.85\textwidth}
\small
\centering
\resizebox{0.95\linewidth}{!}{%
\setlength{\tabcolsep}{2pt}
\renewcommand{\arraystretch}{1.}
\begin{tabular}{c c c c c}
    \toprule
    sampling + classification methods & confusion matrix & recall & precision & $F_1$ \\ \midrule
    RF (original data) &$\left[\begin{array}{cc} 
    693 &6  \\
    35 &7 \end{array}\right]$  &0.17 &0.54 &0.25\\
    NN (original data) &$\left[\begin{array}{cc} 
    675 &24  \\
    31 &11 \end{array}\right]$  &0.26 &0.31 &0.29\\ \midrule
    SMOTE + RF &$\left[\begin{array}{cc} 
    \textcolor{red}{661} &\textcolor{red}{38}  \\
    \textcolor{red}{21} &\textcolor{red}{21} \end{array}\right]$  &\textcolor{red}{0.5} &\textcolor{red}{0.36} &\textcolor{red}{0.42}\\
    SMOTE + NN &$\left[\begin{array}{cc} 
    666 &33  \\
    30 &12 \end{array}\right]$  &0.29 &0.27 &0.28\\ \midrule
    ADASYN + RF &$\left[\begin{array}{cc} 
    \textcolor{red}{662} &\textcolor{red}{37}  \\
    \textcolor{red}{20} &\textcolor{red}{22} \end{array}\right]$  &\textcolor{red}{0.52} &\textcolor{red}{0.37} &\textcolor{red}{0.44}\\
    ADASYN + NN &$\left[\begin{array}{cc} 
    664 &35  \\
    32 &10 \end{array}\right]$  &0.24 &0.22 &0.23\\ \midrule
    Score-based case 1 ($l$=10, $\gamma$=0.2, $\epsilon$=0.01) \\ Score + RF  &$\left[\begin{array}{cc} 
    \textcolor{red}{660} &\textcolor{red}{39}  \\
    \textcolor{red}{11} &\textcolor{red}{31} \end{array}\right]$  &\textcolor{red}{0.74} &\textcolor{red}{0.44} &\textcolor{red}{0.55}\\
    Score + NN &$\left[\begin{array}{cc} 
    653 &46  \\
    22 &20 \end{array}\right]$  &0.48 &0.30 &0.37\\ \midrule
    Score-based case 2 ($l$=20, $\gamma$=0.9, $\epsilon$=0.01) \\
    Score + RF &$\left[\begin{array}{cc} 
    630 &69  \\
    6 &36 \end{array}\right]$  &\underline{\textbf{0.86}} &0.34 &0.49\\
    Score + NN &$\left[\begin{array}{cc} 
    621 &78  \\
    20 &22 \end{array}\right]$  &0.52 &0.22 &0.31\\ \midrule
    Score-based case 3 ($l$=40, $\gamma$=0.9, $\epsilon$=0.0005) \\
    Score + RF &$\left[\begin{array}{cc} 
    671 &28  \\
    12 &30 \end{array}\right]$  &0.71 &0.52 &\underline{\textbf{0.60}}\\
    Score + NN &$\left[\begin{array}{cc} 
    659 &40  \\
    22 &20 \end{array}\right]$  &0.48 &0.33 &0.39\\
    \bottomrule
\end{tabular}
}
\captionof{table}{Discriminative classifiers performance on test set (synthetic data).}
\label{Tab:synthesized_imbalanced_discriminative}
\end{minipage}
\vspace{0.5cm}

\begin{figure}[ht]
    \centering
    \subfloat[\centering SMOTE]{{\includegraphics[width=0.27\columnwidth]{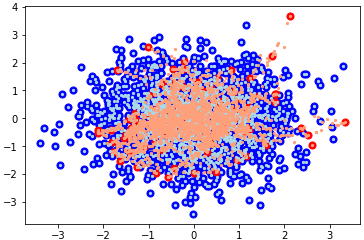}}}
    \quad
    \subfloat[\centering ADASYN]{{\includegraphics[width=0.27\columnwidth]{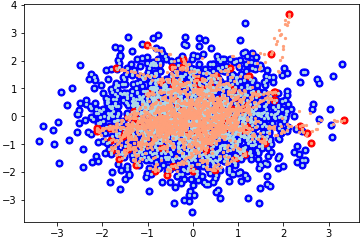}}}
    \\
    \vspace{0.27cm}
    \centering
    \subfloat[\centering Score-based case 1]{{\includegraphics[width=0.27\columnwidth]{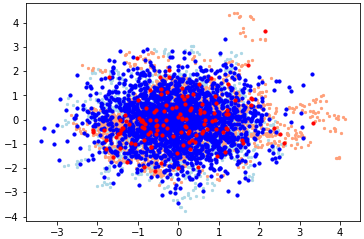}}}
    \subfloat[\centering Score-based case 2]{{\includegraphics[width=0.27\columnwidth]{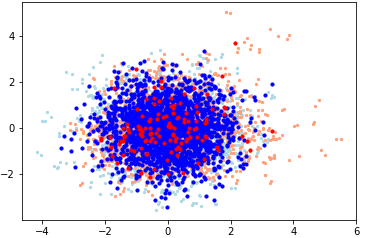}}}
    \subfloat[\centering Score-based case 3]{{\includegraphics[width=0.27\columnwidth]{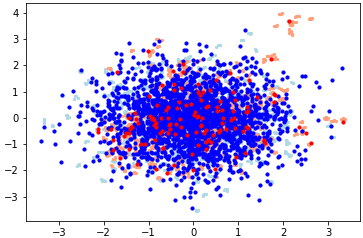}}}
    \\
    \vspace{0.32cm}
    \centering
    \includegraphics[width=0.7\columnwidth]{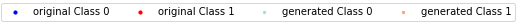}
    \caption{Original and generated samples (showing first two dimensions).}
    \label{fig:imbalanced_classification_compare_new_data}
\end{figure}

We further investigate the logics behind. A comparison of the combined old and new training data generated by the three sampling methods is shown in Fig.\ref{fig:imbalanced_classification_compare_new_data}. We see SMOTE and ADASYN interpolate the existing positive samples, while the learned score function can both interpolate and extrapolate: starting from any existing point, new samples can be generated by choosing proper step size $\epsilon$, walking distance $l$ (i.e. chain length) and discard rate $\gamma$. As these samples are directly sampled from the learned score function (equivalently sampling from the approximate underlying distribution), they are intrinsically representatives of the underlying dynamics. We have thus seen that the score-based generative method is effective in synthesizing rare events in high dimensions.

\subsubsection*{A real-world fraud detection example}

\begin{figure}[ht]
    \centering
    \subfloat[\centering Training]{{\includegraphics[width=0.30\columnwidth]{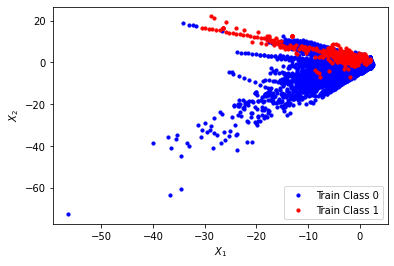}}}
    \qquad
    \subfloat[\centering Test]{{\includegraphics[width=0.30\columnwidth]{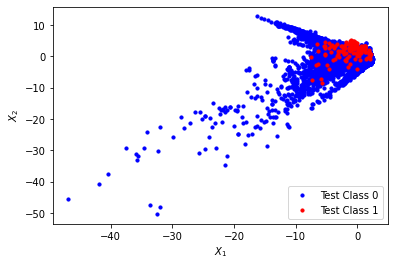}}}
    \caption{Fraud detection data (showing first two dimensions). 18 positives ($\sim$ 5\%) and 18 negatives have their labels flipped.}
    \label{Fig:fraud_detection_2D}
\end{figure}

As our second imbalanced classification example, we analyse an online credit card transaction dataset \cite{credit_card_data, credit_card_Pozzolo, credit_card_Will} which contains credit cards transactions over two days in September 2013 by European cardholders, with 492 frauds out of 284,807 transactions. The negative (non-fraudulent) and positive (fraudulent) samples are highly imbalanced, with a fraud rate of 0.173\%. The original dataset has 30 anonymized features for each transaction record. To demonstrate the efficacy of classification with minimum input indicators, we choose first 10 dimensions and split the dataset into training and test sets with a train to test ratio of about 3:1. Further, to test the robustness of the score-based generative sampling, we randomly swap the labels of a small number ($18 \times 2$) of training examples. Details about the data is presented in Table.\ref{Tab:fraud_detection_data_specification}. The first two dimensions of the training and test sets are visualised in Fig.\ref{Fig:fraud_detection_2D}.

\small
\begin{table}[ht] 
\begin{center}
\scalebox{0.85}{
\begin{tabular}{c c c c} 
    \toprule
    Set\textbackslash description & size & No.positive & No.negative \\ \midrule
    Training &212332 &369 &211963 \\ \midrule
    Test &72475 &123 &72352 \\ \midrule
    Total &284807 &492 &284315 \\ \bottomrule
\end{tabular}}
\caption{Specification of fraud detection data.} \label{Tab:fraud_detection_data_specification}
\end{center}
\end{table}
\normalsize

\normalsize

Without further feature engineering (e.g. normalization), we build a plain XGB booster, one of the practically popular and computationally efficient emsemble classifiers, and results shown in Table.\ref{Tab:fraud_detection}. It is seen that, without data augmentation, the classifier correctly identifies 54\% (67 out of 123) of all the positive cases in the test set, making in total 58 mistakes (2 FPs and 56 FNs). We then separately apply SMOTE, ADASYN and the score-based methods to upsample the minority classes, augmenting the number of positive samples to be 10 folds (i.e. 3690 cases) of the original cases and achieving a constant positive-to-negative ratio of about 1.74:100 (more samples can be generated if computational resource allows). For score-based sampling, a score function is trained using the 369 positive samples (we only need to learn one score function for the minor class), and then used to generate new positive samples which are added to the original training data.

\begin{figure}[ht] 
    \centering
    \subfloat[\centering SMOTE]{{\includegraphics[width=0.27\columnwidth]{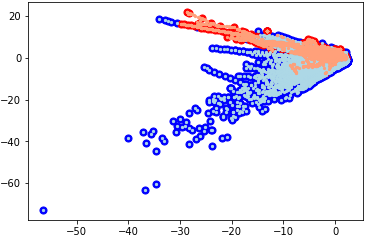}}}
    \quad
    \subfloat[\centering ADASYN]{{\includegraphics[width=0.27\columnwidth]{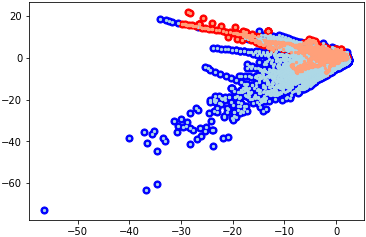}}}
    \\
    \vspace{0.27cm}
    \centering
    \subfloat[\centering Score-based case 1]{{\includegraphics[width=0.27\columnwidth]{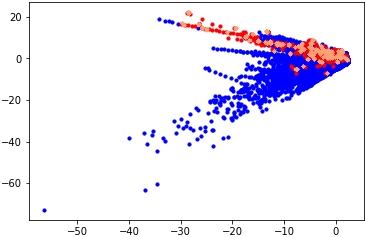}}}
    \quad
    \subfloat[\centering Score-based case 2]{{\includegraphics[width=0.27\columnwidth]{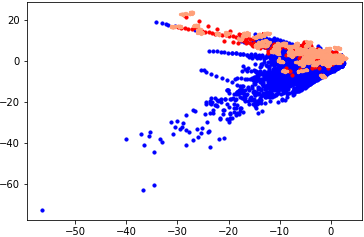}}}
    \quad
    \subfloat[\centering Score-based case 3]{{\includegraphics[width=0.27\columnwidth]{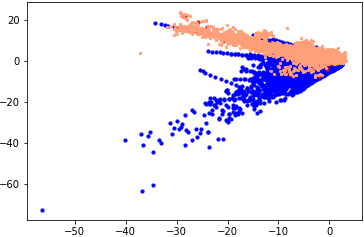}}}
    \\
    \vspace{0.32cm}
    \centering
    \includegraphics[width=0.7\columnwidth]{fig/data_generation_label.png}
    \caption{Original and generated samples (showing first two dimensions). All sampling methods produce about 3690 (10x original) positive samples, contrasting 211963 negative samples (positive-to-negative ratio 1.74:100).}
    \label{fig:fraud_detection_sampling}
\end{figure}

When sampling using the learned score function, there are Langevin hyperparameters ($l, \gamma, \epsilon$) which can be chosen flexibly. Each set of hyper-parameter randomly generates a set of samples, as shown in Fig.\ref{fig:fraud_detection_sampling}, and corresponding results in Table.\ref{Tab:fraud_detection}. Notable observations are: first, comparing the performances using original data and score-generated data (see case 1), the later yields better performance across all metrics. Second, the SMOTE and ADASYN generated data improve the classifier's ability in identifying positives, but also introduce more FPs. Third, score-generated data show superior performance over SMOTE and ADASYN generated data (e.g. see case 2), which demonstrates the effectiveness of score-based generative sampling. Fourth, score-generated data yield comparable or better performance to all other methods in terms of total mistakes made. Considering noise introduced in the training set (36 flipped labels), the results support the argument that, score-based generative modelling could be more robust to perturbations or outliers. By generating more credible samples, it can migrate the disturbing effects of adversarial samples.

\vspace{0.5cm}
\begin{center}
\begin{minipage}{0.9\textwidth}
\small
\centering
\resizebox{0.95\linewidth}{!}{%
\setlength{\tabcolsep}{2pt}
\renewcommand{\arraystretch}{1.}
\begin{tabular}{c c c c c c}
    \toprule
    sampling + classification methods & confusion matrix & recall & precision & $F_1$ & FPs+FNs \\ \midrule
    XGB (original data) &$\left[\begin{array}{cc} 
    \textcolor{red}{72350} &\textcolor{red}{2}  \\
    \textcolor{red}{56} &\textcolor{red}{67} \end{array}\right]$  &\textcolor{red}{0.54} &\textcolor{red}{0.97} &\textcolor{red}{0.70} &\textcolor{red}{58}\\ \midrule
    SMOTE + XGB &$\left[\begin{array}{cc} 
    72341 &11  \\
    39 &84 \end{array}\right]$  &0.68 &0.88 &0.77 &50\\ \midrule
    ADASYN + XGB &$\left[\begin{array}{cc} 
    72339 &13  \\
    38 &85 \end{array}\right]$  &0.69 &0.87 &0.77 &51\\ \midrule
    Score-based case 1 ($l$=300, $\gamma$=0.1, $\epsilon$=0.0001) \\
    Score + XGB &$\left[\begin{array}{cc} 
    \textcolor{red}{72350} &\textcolor{red}{2}  \\
    \textcolor{red}{43} &\textcolor{red}{80} \end{array}\right]$  &\textcolor{red}{0.65} &\textcolor{red}{0.98} &\textcolor{red}{0.78} &\textcolor{red}{45}\\
    Score-based case 2 ($l$=100, $\gamma$=0.3, $\epsilon$=0.01) \\
    Score + XGB &$\left[\begin{array}{cc} 
    72342 &10  \\
    34 &89 \end{array}\right]$  &0.72 &0.90 &\underline{\textbf{0.80}} &\underline{\textbf{44}}\\
    Score-based case 3 ($l$=10, $\gamma$=0.1, $\epsilon$=0.05) \\
    Score + XGB &$\left[\begin{array}{cc} 
    72332 &20  \\
    32 &91 \end{array}\right]$  &\underline{\textbf{0.74}} &0.82 &0.78 &52\\ \bottomrule
\end{tabular}
}
\captionof{table}{Discriminative classifiers performance on test set (fraud detection).}  \label{Tab:fraud_detection}
\end{minipage}
\end{center}
\vspace{0.5cm}

\section{Score-based generative classification}
\label{Sec:score-based generative classification}

\paragraph{Density representation and decision theory}

In generative modelling, we focus on $p(x|y)$ (or just $p(x)$ when we are referring to a particular class). We have seen examples of Gaussian scores in Fig.\ref{Fig:1D_normal_density_and_score_func} and Fig.\ref{Fig:2D_normal_density_and_score_func}. If we have multiple DGPs, each class-conditional density $p(x|y_j)$ can be modelled as $p(x|y_j) = \frac{e^{-f_{\theta_j}(x)}}{Z_{\theta_j}}$. Two classes as an example, their densities can be written as $p_{\theta_0}(x)=\frac{e^{-f_{\theta_0}(x)}}{Z_{\theta_0}}$ and $p_{\theta_1}(x)=\frac{e^{-f_{\theta_1}(x)}}{Z_{\theta_1}}$, and corresponding score functions derived as $s_{\theta_0}=-\nabla_x f_{\theta_0}(x), s_{\theta_1}=-\nabla_x f_{\theta_1}(x)$. Unlike the discriminative density $p(y|x)$ with property $p_{\theta_0}(y=0|x)+p_{\theta_1}(y=1|x)=1$, which leads to a shared normalising constant, the constants $Z_{\theta}$ in the generative densities $p(x|y)$ don't naturally connect to each other, neither are they necessarily the same.

If we encode equal prior probabilities for both classes, the Bayes rule (Eq.\ref{Eq:generative_classification_1}) can be simplified: 

\begin{equation} \label{Eq:generative_classification_2} \tag{\ref{Eq:generative_classification_1}b}
    p(y_j|x) = \frac{p(x|y_j)}{\sum_{j=1}^{c} p(x|y_j)}
\end{equation}

\noindent and the decision rule reduces to simply choosing the class label with higher density value, which again gives the \textit{Bayes classifier}. Binary classification, for example, has the generative decision rule (using a \textit{soft margin} $\gamma_0=0$ in accordance with the discriminative rule Eq.\ref{Eq:binary_decision_theory_2}): 

\begin{equation} \label{Eq:binary_generative_decision_theory}
\hat{y} = 
\begin{cases}
    1, & \text{if } \log \frac{p(x|y=1)}{p(x|y=0)} \geq 0 \\
    0,              & \text{otherwise}
\end{cases}
\end{equation}

This simple density contrasting rule has previously been exercised in Fig.\ref{fig:Gaussian_pdfs_recovered_by_scores}. To apply this rule, a key step is to construct the class-conditional density $p_{\theta}(x)$ from score function $s_{\theta}(x)$. This can be done using the density construction method described in Section.\ref{Sec:score_based_generative_modelling} (e.g. Algorithm.\ref{algo:reconstruct_pdf_from_score_func}), which requires an initial probability $p(x_0)$, normally empirically estimated, to be supplied. If we have additional knowledge about the samples, e.g. a cluster is Gaussian distributed, we can make use of this information in constructing the density, and potentially obtain better density estimation and decisions. We describe this using the 1D and 2D Gaussian data examples in the following.

An alternative and costive approach is to numerically solve the decision boundary equation in Eq.\ref{Eq:binary_generative_decision_theory} via \textit{NR} which makes use of score function. Using the above specified density representations, boundary condition can be written as:

\begin{equation} \label{Eq:d_x_generative}
    \gamma(x) = \log \frac{p(x|y=1)}{p(x|y=0)} = f_{\theta_0}(x) - f_{\theta_1}(x) + \log Z_{\theta_0} - \log Z_{\theta_1}
    = 0 
\end{equation}

\noindent To numerically find its roots (i.e. points lying on the boundary), the \textit{NR} updating formula (Eq.\ref{Eq:Newton_Raphson}) is applied, in which the derivative $\nabla_x \gamma(x) = \nabla_x f_{\theta_0}(x) - \nabla_x f_{\theta_1}(x) = s_{\theta_1}(x) - s_{\theta_0}(x)$. However, when using Eq.\ref{Eq:Newton_Raphson} updates, we are blocked by evaluating $\gamma(x)$, which involves assessing $\log Z_\theta$ that could be intractable. If we make some distributional assumption about data, e.g. each class is Gaussian distributed, then we can empirically estimate $\log Z_\theta$ (and other constants associated with the assumed density). However, this raises the question: if we know the distributional properties of the clusters, we could directly construct their densities using estimated sufficient statistics and draw the decision boundary already, so why bother learning their score functions, performing integration and/or going numerical with extra costs? There are two reasons: first, learning a score function in many cases is cheap and fast, almost marginal with aid of modern ML techniques such as deep learning and automatic differentiation (both are used in score matching, e.g. a shallow multi-layer perceptron can achieve satisfying loss). Second, we will see from following Gaussian data examples that, the decision boundary found by \textit{NR} via scores are very close to theoretical ones, despite the inaccuracies in score learning. 

\subsection{Two toy Gaussian data examples}

We apply generative classification to the simulated 1D (Fig.\ref{Fig:1D_normal_density_and_score_func}) and 2D (Fig.\ref{fig:2D_Gaussian_DGP}) Gaussian data. Given the fact that there are equal number of positives and negatives in both scenarios (1000 for each class in the 1D Gaussian case, and 200 in the 2D case), the prior probabilities for both classes are equal, and the decision rule Eq.\ref{Eq:generative_classification_2} can be directly applied.

Pretending that we have no access to DGPs, we just assume Gaussian distribution for both clusters (by visually inspecting the data), which hints the theoretical normalising constant $Z=(2\pi)^{d/2}|\Sigma|^{1/2}$ (see Gaussian density in Eq.\ref{Eq:MND_pdf}), and score function $s(x|\mu,\Sigma)=\Sigma^{-1}(\mu-x)$. We learn two parametric, linear score functions $\hat{s}_\theta(x)=\hat{A}x+\hat{b}$ (here $\theta=[\hat{A}, \hat{b}]$) via score matching; these learned score functions can be analytically integrated to yield $\hat{f}_\theta(x)=-\frac{1}{2}x^T\hat{A}x-\hat{b}^Tx+\hat{C}$ (Eq.\ref{Eq:score_function_definition_2}), where the constant $\hat{C}$ can be obtained by using $\hat{f}_\theta(\hat{\mu})=0$ (Gaussian property):

\begin{equation} \label{Eq:estimated_Gaussian_f}
    \hat{f}_\theta(x) = -\frac{1}{2}x^T\hat{A}x-\hat{b}^Tx+\frac{1}{2}{\hat{\mu}}^T\hat{A}{\hat{\mu}}+\hat{b}^T{\hat{\mu}}
\end{equation}

\noindent where $\hat{\mu}$ is the sample mean. We can also derive a sample-based guess $\hat{Z_{\theta}}=(2\pi)^{d/2}|\hat{\Sigma}|^{1/2}$, which enables construction of an estimated density:

\begin{equation} \label{Eq:estimated_generative_density}
    \hat{p}_\theta(x) = e^{-\hat{f}_{\theta}(x)} / \hat{Z}_\theta
\end{equation}

\noindent Note Eq.\ref{Eq:estimated_generative_density} has been derived using the strong assumption of distributional structure of data.

Alternatively, given $\hat{s}_\theta(x)$, $\hat{f}_\theta(x)$ and $\hat{Z}_\theta$, we can numerically find the boundary using the \textit{NR} updating rule (Eq.\ref{Eq:Newton_Raphson}) which solves the boundary equation Eq.\ref{Eq:d_x_generative}:

\begin{equation} \label{Eq:Newton_Raphson_generative_classification}
    x_{k+1} = x_k - \frac{\hat{f}_{\theta_0}(x) - \hat{f}_{\theta_1}(x) + \log \hat{Z}_{\theta_0} - \log \hat{Z}_{\theta_1}}{\hat{s}_{\theta_1}(x) - \hat{s}_{\theta_0}(x)}
\end{equation}

We have therefore two equivalent choices at hand in seeking the boundary: either directly comparing $e^{-\hat{f}_{\theta_0}(x)}/\hat{Z}_{\theta_0}$ and $e^{-\hat{f}_{\theta_1}(x)}/\hat{Z}_{\theta_1}$ at grid points with assistance of Eq.\ref{Eq:estimated_Gaussian_f}, or using a numerical scheme such as \textit{NR}. We test both methods for the 1D (Fig.\ref{Fig:1D_normal_density_and_score_func}) and 2D (Fig.\ref{fig:2D_Gaussian_DGP}) Gaussian data.

\begin{figure}[ht] 
    \centering
    \subfloat[\centering Class 0 (JSD: $\sim 0$)]{{\includegraphics[width=0.35\columnwidth]{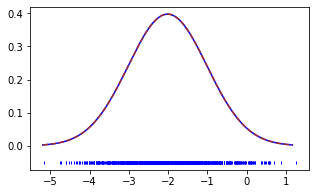}}}
    \qquad
    \subfloat[\centering Class 1 (JSD: $\sim 0$)]{{\includegraphics[width=0.35\columnwidth]{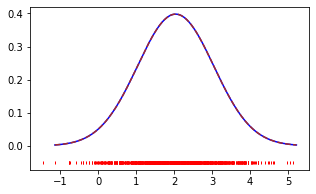}}}
    \\
    \includegraphics[width=0.34\columnwidth]{fig/two_1D_Gaussian_scores_legends.png}
    \caption{Recovered densities using learned (analytical) score function. There are 1000 samples for each class, with locations shown at bottom. The score functions use a two-layer MLP with sizes [1,1], representing a linear function $\hat{s}(x)=\hat{a}x+\hat{b}$. Initial probability $p_{\theta_0}(\hat{\mu}_0=[-2.01])=0.40,p_{\theta_1}(\hat{\mu}_1=[2.04])=0.40$.}
    \label{fig:1D_Gaussian_pdfs_recovered_from_score_func_assumeGaussian}
\end{figure}

We first present the toy 1D Gaussian results. The sample means and standard deviations are $\hat{\mu}_0=-2.01,\hat{\sigma}_0=1.0,\hat{\mu}_1=2.04,\hat{\sigma}_1=1.0,$ The learned score functions, optimized via score matching, are $\hat{s}_{\theta_0}(x)=-0.99x-2$ and $\hat{s}_{\theta_1}(x)=-0.99x+2$. They are close to the theoretical score function $s(x)=(\mu-x)/\sigma$ with DGP parameters $\mu=\pm 2, \sigma=1$. Using Eq.\ref{Eq:estimated_Gaussian_f} and Eq.\ref{Eq:estimated_generative_density}, the two densities, $\hat{p}_{\theta_0}(x)$ and $\hat{p}_{\theta_1}(x)$, are recovered in Fig.\ref{fig:1D_Gaussian_pdfs_recovered_from_score_func_assumeGaussian}. We observe improvement over Fig.\ref{fig:1D_Gaussian_pdfs_recovered_from_score_func}: with the additional distributional assumption, we are able to coin analytical score functions and evaluate density constants from samples, which reduces uncertainties and errors, and rewards a higher recovery accuracy, evidenced by near-zero JSD in both cases. 

Direct comparison of the two estimated densities $\hat{p}_{\theta_0}(x)$ and $\hat{p}_{\theta_1}(x)$ gives the decision boundary around $x=0$. \textit{Newton-Raphson} finds the decision boundary at $x=0.017$. Both are close to the density cross-over point $x=0$ where the generative decision rule (Eq.\ref{Eq:generative_classification_2}) points to.

For the 2D case, sample means and covariances are $\hat{\mu}_0=[0.07, -0.12], \hat{\mu}_1=[4.06, 4.18]$, 
$\hat{\Sigma}_0 =
\begin{bmatrix}
1.08 & -0.56 \\
-0.56 & 1.02
\end{bmatrix}
$,
$\hat{\Sigma}_1 =
\begin{bmatrix}
1.01 & 0.56 \\
0.56 & 1.06
\end{bmatrix}
$,
The learned score functions $\hat{s}(x)=\hat{A}x+\hat{b}$ have coefficients $\hat{A}_0 =
\begin{bmatrix}
-1.30 & -0.71 \\
-0.71 & -1.38
\end{bmatrix}
$,
$\hat{b}_0=(0.01,-0.11)$
,
$\hat{A}_1 =
\begin{bmatrix}
-1.41 & 0.75 \\
0.75 & -1.34
\end{bmatrix}    
$
, 
$\hat{b}_1=(2.57,2.57)$.
The learning is imperfect though: theoretically we expect $A_0=-\Sigma_0^{-1}=\begin{bmatrix}
-1.33 & -0.67 \\
-0.67 & -1.33
\end{bmatrix}, b_0=\Sigma_0^{-1}\mu_0=[0, 0]$ and $A_1=-\Sigma_1^{-1}=\begin{bmatrix}
-1.33 & 0.67 \\
0.67 & -1.33
\end{bmatrix}, b_1=\Sigma_1^{-1}\mu_1=[2.67, 2.67]$. The errors might be induced by finite samples, e.g. the estimated sample mean $\hat{\mu}_0=[0.07, -0.12]$ deviates from true DGP parameter $\mu_0=[0,0]$, \textit{etc}.

\begin{figure}[ht] 
    \centering
    \subfloat[\centering Score-recovered densities]{{\includegraphics[width=0.43\columnwidth]{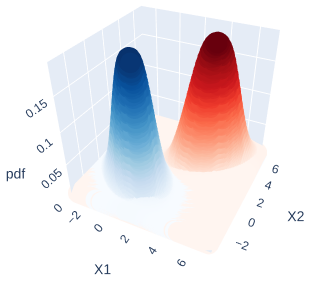}}}
    \qquad
    \subfloat[\centering 2D projections]{{\includegraphics[width=0.35\columnwidth]{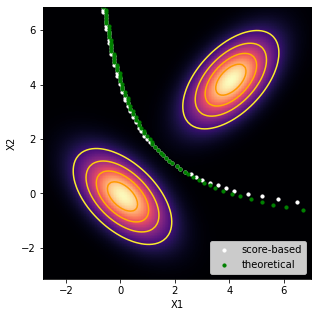}}}
    \caption{Recovered densities and decision boundary using learned (analytical) score function. There are 200 samples for each class. The score functions use a two-layer MLP with sizes [2,2], representing a linear function $\hat{s}(x)=\hat{A}x+\hat{b}$. Initial probability $p_{\theta_0}(\hat{\mu}_0=[0.07,-0.12])=0.18, p_{\theta_1}(\hat{\mu}_1=[4.06,4.18])=0.18$. On the right, brighter color implies higher recovered density, contour and green dots represent DGP-based (theoretical), white dots indicate the equal density boundary generated by recovered densities.}
    \label{fig:Gaussian_pdfs_recovered_by_scores_assumeGaussian}
\end{figure}

The estimated densities from Eq.\ref{Eq:estimated_generative_density} and resulted decision boundary are presented in Fig.\ref{fig:Gaussian_pdfs_recovered_by_scores_assumeGaussian}. Compared with Fig.\ref{fig:Gaussian_pdfs_recovered_by_scores}, the analytically recovered densities and boundary are of better quality (i.e. thinner and less heavy tails), and closer to the theoretical ones in Fig.\ref{fig:2D_Gaussian_DGP}(a). This is due to the extra assumption of Gaussian distribution of data, which gives linear form of score function and leads to (approximately) accurate estimation of the constants in the density (i.e. $\hat{C}$ in Eq.\ref{Eq:estimated_Gaussian_f} and $\hat{Z_{\theta}}$ in Eq.\ref{Eq:estimated_generative_density}). 

\begin{figure}
    \centering
    \includegraphics[width=0.33\linewidth]{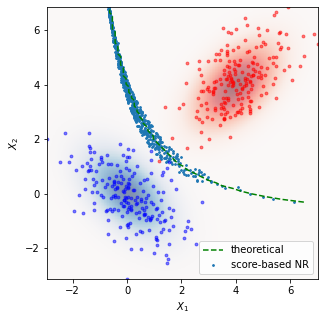}
    \caption{Theoretical and score-based generative decision boundaries for the 2D Gaussian data from Fig.\ref{fig:2D_Gaussian_DGP}.}
    \label{fig:2D_Gaussian_generative_boundary}
\end{figure}

The decision boundary found by the \textit{NR} method are presented in Fig.\ref{fig:2D_Gaussian_generative_boundary}. This numerically derived boundary (blue dots) trembles around the theoretical one (green dots). Despite inaccuracies exist in score function learning, the resulted boundaries are still satisfying, which could be (hesitantly) attributed to the denominator and numerator cancellation effect in the \textit{NR} updating formula. An extended theoretical treatise of the boundary between two Gaussian densities can be found in Appendix.\ref{App:Gaussian_scores_and_separality}.  

Compared to the density construction method in Section.\ref{Sec:score_based_generative_modelling}, which requires only the learned score functions $\hat{s}_\theta(x)$ and initial point probabilities $\hat{p}_\theta(x_0)$, the method used here provides accurate boundary inference. However, when distributional assumption cannot be made, or the learned score function is complex (e.g. deep NN) and cannot be analytically integrated, we may have to stick to the former method. This is illustrated in following two examples.

\subsection{Generative classification of imbalanced data}

The two imbalanced datasets, i.e. the medium-size, high-dimensional synthetic data, and the large-volume, real-world fraud detection data, both used in discriminative classification, are again employed. As the two classes are highly imbalanced (positive-to-negative ratio of 1:95 in the synthetic case, and fraud rate of 0.173\% in the second case), considering data imbalance, the prior probabilities could exert great influence on the posterior, we therefore adopt the original Bayes rule (Eq.\ref{Eq:generative_classification_1}).

As we are contrasting the two posteriors, we need to learn a score function for each class. However, this time we won't try to construct the whole density surface and find the boundary; instead we query the score function (the 'oracle'), along with a supplied initial density value, on a come-and-serve basis. The initial density values can be estimated using two methods: the neighbourhood counting method which divides frequencies of class appearance in a $\sigma$-neighbourhood (with radius $\sigma$) of $x_0$ by its total number and volume, and the Gaussian estimation method which gives $\hat{p}(x_0=\hat{\mu})=1/(2\pi)^{d/2}|\hat{\Sigma}|^{1/2}$. The former gives rough estimation, while the later makes a weak, partial Gaussian assumption on data. Unlike the strong Gaussian distributional assumption used in the 1D and 2D Gaussian data examples, the linear structure assumption are not imposed in score function learning, we are free to choose any proper neural network architecture (not limited to two-layer and linear) to fit the gradient field via score matching. However, still lack of evidence of Gaussianality, e.g. density peaks at sample mean, could induce inaccurate score estimation, which is particularly unfair for multimodal data. In the simulated imbalanced data example, we use the Gaussian initial density estimates; we compare both methods in the real-world fradu detection example.

\subsubsection{The simulated high-dimensional imbalanced data example}

This time we use different settings for learning the score functions: a slightly shallower neural network architecture with layer sizes $[10,128,128,10]$ is used for both score functions, demonstrating the flexibility of score function representation. We also use different learning rates to update weights during backpropagation, $\alpha_0=0.001$ for the major class (labeled 0), and $\alpha_1=0.01$ or $0.05$ for the minor class (labeled 1), allowing for different levels of details to be learned adapting to data volume. 

The empirical prior probabilities for the major and minor classes are 0.943 and 0.057, given by the two class ratios. An (rough) visual inspection of the first two dimensions of the data (Fig.\ref{Fig:imbalanced_simulated_data}) hints that we can use a Gaussian guess of the initial density, which gives $\hat{p}_{\theta_0}(x_0=\hat{\mu}_0)=0.9 \times 10^{-4}$ and $\hat{p}_{\theta_1}(x_0=\hat{\mu}_1)=1.2 \times 10^{-4}$. The two densities $p_{\theta_0}(x)$ and $p_{\theta_1}(x)$ at each test point are calculated incrementally using segment line integral (Eq.\ref{pdf_reconstruction_from_score_func_2D_integral}) with the supplied initial density values. The posterior, computed by Eq.\ref{Eq:generative_classification_1}, are compared and data classified as the class with higher posterior. The results yielded by two learned score models are reported in Table.\ref{Tab:synthesized_imbalanced_generative}. The generative classifiers yield comparable (marginally better) performance to the SMOTE and ADASYN based discriminative classifiers in Table.\ref{Tab:synthesized_imbalanced_discriminative}: classifier 1 identifies the same number of positives (21 TPs out of 42 overall positives) as the random forest (RF) discriminative classifier with SMOTE upsampling, with less FPs (27 \textit{vs} 38); classifier 2 also shows similar performance (22 TPs correctly identified, 35 FPs made by generative classifier 2 \textit{vs} 37 FPs made by RF + ADASYN). Compared to the score-assisted discriminative classifiers (last 3 rows in Table.\ref{Tab:synthesized_imbalanced_discriminative}), the generative classifiers show inferior performance to the random forest classifiers using score-augmented data, but superior to the score-assisted neural network classifier. 

\vspace{0.5cm}
\begin{center}
\begin{minipage}{1.0\textwidth}
\small
\centering
\resizebox{0.95\linewidth}{!}{%
\setlength{\tabcolsep}{2pt}
\renewcommand{\arraystretch}{1.}
\begin{tabular}{c c c c c}
    \toprule
    Scenario & confusion matrix & recall & precision & $F_1$ \\ \midrule
    Score-based generative classifier 1 ($\alpha_0=0.001, \alpha_1=0.01$) &$\left[\begin{array}{cc} 
    \textcolor{black}{672} &\textcolor{black}{27}  \\
    \textcolor{black}{21} &\textcolor{black}{21} \end{array}\right]$  &\textcolor{black}{0.50} &\textcolor{black}{0.44} &\underline{\textcolor{black}{\textbf{0.47}}} \\
    Score-based generative classifier 2 ($\alpha_0=0.001, \alpha_1=0.05$) &$\left[\begin{array}{cc} 
    \textcolor{black}{664} &\textcolor{black}{35}  \\
    \textcolor{black}{20} &\textcolor{black}{22} \end{array}\right]$  &\underline{\textcolor{black}{\textbf{0.52}}} &\textcolor{black}{0.39} &\textcolor{black}{0.44} \\
    \bottomrule
\end{tabular}
}
\captionof{table}{Generative classifiers performance on test set (synthetic data, $\alpha$: learning rate of score network).}  \label{Tab:synthesized_imbalanced_generative}
\end{minipage}
\end{center}
\vspace{0.5cm}

\subsubsection{The fraud detection example}

We apply generative classification to the highly imbalanced, ten-dimensional fraud detection dataset (with in total 36 flipped class labels). Unlike the discriminative case where we only learn a score function for the minor class to augment data, this time we need to learn separately two score functions and use them to predict class-specific probabilities for each test point. Again we built two generative classifiers with different learning settings, and report the results in Table.\ref{Tab:fraud_detection_generative}. Features are scaled using z-score standardization before training. The same neural network architecture, with layer sizes [10, 128, 128, 10], are used for representing score functions for both classes. In training the score functions, stochastic gradient descent (SGD) with batch sizes of 128 (major class) and 64 (minor class) is used to accelerate the training process. Note that, in contrast to the case of synthetic example, we apply smaller learning rate for the minor class with sufficiently large number of epochs, which allows the score network to learn more details from the data while evolving slowly.

Two initial density estimation methods are used: the neighbourhood counting method and the Gaussian density estimation method. Both methods estimate an initial density value $\hat{p}_\theta(x_0=\hat{\mu})$ at the sample mean location. A visual inspection of the first two dimensions (Fig.\ref{Fig:fraud_detection_2D}) suggests that the data may not be Gaussian distributed, which invalidates the Gaussian initial density method. The $\sigma$-neighbourhood counting method calculates the ratio of samples within a circle of radius $\sigma=1.0$ centering $x_0=\hat{\mu}$ (i.e. the area under \textit{pdf} in the neighbourhood of $x_0$). These two methods may give different estimation values, however, as we are contrasting the two densities, the absolute values may be irrelevant once they give similar scale class density ratios. The Gaussian method yields an initial density ratio $\hat{p}_{\theta_0}(x_0)/\hat{p}_{\theta_1}(x_0)$ of 140 while the counting method gives $\sim 70$. It is observed that, the neighbourhood counting method gives slightly better results in terms of recall (and similar $F_1$ values), at the expense of worse precision (i.e. more FPs).

A fair comparison can be made between the test performances of the generative classifier (Table.\ref{Tab:fraud_detection_generative}) and the discriminative XGB classifier (first row in Table.\ref{Tab:fraud_detection}), both built on original data. The generative classifiers identify marginally more positives (indicated by higher recalls) at the expense of introducing more FPs. We are cautious to compare the generative results to other data-augmented discriminative cases, neither can it be compared to other state-of-the-art hand crafted methods, as the data (only 10 features features and original samples are used), feature engineering (only standardization is used), and noise levels (36 labels swapped) used could be different.

\vspace{0.5cm}
\begin{center}
\begin{minipage}{1.0\textwidth}
\small
\centering
\resizebox{1.0\linewidth}{!}{%
\setlength{\tabcolsep}{2pt}
\renewcommand{\arraystretch}{1.1}
\newcolumntype{P}[1]{>{\centering\arraybackslash}p{#1}}
\newcolumntype{M}[1]{>{\centering\arraybackslash}m{#1}}
\begin{tabular}{c c c c c c}
    \toprule
    $p(x_0)$ estimation method & Scenario & confusion matrix & recall & precision & $F_1$ \\ \midrule
    \multirow{3}{*}{\shortstack[l]{Gaussian \\ $\hat{p}_{\theta_0}(x_0=\hat{\mu}_0)=7.0 \times 10^{-6}$\\ $\hat{p}_{\theta_1}(x_0=\hat{\mu}_1)=5.0 \times 10^{-8}$}}
        & \shortstack[l]{Score-based generative classifier 1 \\($\alpha_0=0.05, \alpha_1=0.001$)} &$\left[\begin{array}{cc} 
        \textcolor{black}{72166} &\textcolor{black}{186}  \\ 
        \textcolor{black}{51} &\textcolor{black}{72} \end{array}\right]$  &\underline{\textcolor{black}{\textbf{0.59}}} &\textcolor{black}{0.28} &\textcolor{black}{0.38} \\ \cline{2-6}
        & \shortstack[l]{Score-based generative classifier 2 \\($\alpha_0=0.05, \alpha_1=0.0005$)} &$\left[\begin{array}{cc} 
        \textcolor{black}{72269} &\textcolor{black}{83}  \\
        \textcolor{black}{52} &\textcolor{black}{71} \end{array}\right]$  &\textcolor{black}{0.58} &\textcolor{black}{0.46} &\textcolor{black}{0.51} \\ \cline{2-6}
        & \shortstack[l]{Score-based generative classifier 3 \\($\alpha_0=0.01, \alpha_1=0.0005$)} &$\left[\begin{array}{cc} 
        \textcolor{black}{72299} &\textcolor{black}{53}  \\
        \textcolor{black}{59} &\textcolor{black}{64} \end{array}\right]$  &\textcolor{black}{0.52} &\textcolor{black}{0.55} &\underline{\textbf{\textcolor{black}{0.53}}} \\ \cline{1-6}
        
        \multirow{3}{*}{\shortstack[l]{Neighbourhood counting \\ $\hat{p}_{\theta_0}(x_0=\hat{\mu}_0)=0.548$\\ $\hat{p}_{\theta_1}(x_0=\hat{\mu}_1)=0.008$}}
         & \shortstack[l]{Score-based generative classifier 1 \\($\alpha_0=0.05, \alpha_1=0.001$)} &$\left[\begin{array}{cc} 
        \textcolor{black}{72148} &\textcolor{black}{204}  \\ 
        \textcolor{black}{47} &\textcolor{black}{76} \end{array}\right]$  &\underline{\textcolor{black}{\textbf{0.62}}} &\textcolor{black}{0.27} &\textcolor{black}{0.38} \\ \cline{2-6}
        & \shortstack[l]{Score-based generative classifier 1 \\($\alpha_0=0.05, \alpha_1=0.0005$)} &$\left[\begin{array}{cc} 
        \textcolor{black}{72261} &\textcolor{black}{91}  \\ 
        \textcolor{black}{48} &\textcolor{black}{75} \end{array}\right]$  &\textcolor{black}{0.61} &\textcolor{black}{0.45} &\textcolor{black}{0.52} \\ \cline{2-6}
        & \shortstack[l]{Score-based generative classifier 2 \\($\alpha_0=0.01, \alpha_1=0.0005$)} &$\left[\begin{array}{cc} 
        \textcolor{black}{72285} &\textcolor{black}{67}  \\
        \textcolor{black}{55} &\textcolor{black}{68} \end{array}\right]$  &\textcolor{black}{0.55} &\textcolor{black}{0.50} &\underline{\textcolor{black}{\textbf{0.53}}} \\ 
    \bottomrule
\end{tabular}
}
\captionof{table}{Generative classifiers performance on test set (fraud detection, $\alpha$: learning rate of score network).} \label{Tab:fraud_detection_generative}
\end{minipage}
\end{center}
\vspace{0.5cm}

\section{Discussions}

In this work, we investigated the applications of score-based generative modelling in discriminative and generative classification settings. Specifically,

\begin{itemize}
    \item We illustrated the fundamental principles of score-based generative modelling, including score function as an alternative to density for characterizing data distribution, learning individual and generic score functions via score matching, constructing densities from scores in low and high dimensions, and sample generation via Langevin dynamics. 
    \item We discussed the basic decision theories for both discriminative and generative classification problems, and how score functions can be involved in making decisions. For generative classification, learned score functions can be used to construct class-conditional densities which is used directly in the Bayes rule; for discriminative classification, it can be used to augment sample space, or to characterize learned densities in the post-classification stage. In both tasks, scores can be used to find the decision boundary with \textit{Newton-Raphson} method.
    \item We performed discriminative and generative classification experiments on three datasets: a simulated 1D and 2D Gaussian dataset of small size, a simulated imbalanced dataset of medium size and high dimension, and a large volume, high-dimensional, highly imbalanced real-world fraud detection dataset. The score-based data augmentation methodology was benchmarked by SMOTE and ADASYN oversampling methods, and exhibited superior performance in imbalanced classification tasks. The score-based generative classifiers yield comparable performance to discriminative classifiers.
\end{itemize}

Across this work, these techniques are explained with analytical, numerical and experimental examples. Various classifiers are applied in different settings, i.e. nearest neighbours, logistic regression, random forest, extreme gradient booster, neural network for discriminative classification, and (implicitly) Bayes classifier for generative classification. Gaussian densities are employed in many of our analytical and simulation examples, due to its analytical tractability and numerical verifiability. To foster further discussions, we address some interesting issues identified in our work.

\paragraph{Learning the score network}

We can write a score function in parametric form (e.g. Gaussian scores), or represent it using a neural network, which is of advantage as neural network can model complex, high-dimensional, non-linear relations. Learning a score function, in many cases, is cheap and fast (e.g. optimization via first-order methods such as gradient descent). Also, training a small to medium size neural network is made fast with aid of modern learning techniques (e.g. SGD). At inference time, however, it could be slow if the size of test samples is large (as in the fraud detection example). The computational intensity gets worse if Monte Carlo integral is evaluated incrementally when inferring densities.

When learning a score network, it might be beneficial to standardize or normalize the input features for two reasons: first, to assign (roughly) equal weight to each feature, because features with large values could exert more influence on loss. Second, to accelerate the learning process. Standardization can help reshape the skewed loss surface to an standard ellipse, which facilitates gradient descent updating. We exercised this in generative classification of fraud detection data, but have not benchmarked it. With or without feature scaling might lead to different learned score fields, and therefore influence subsequent density construction and classification outcomes. Also, in the presence of large volume data (e.g. the major class in fraud detection), stochastic gradient descent can be used for training the score network, although bias and inaccuracies may be introduced. 

\paragraph{Trustful data augmentation} 

Score function can be used to generate synthetic data with high fidelity.
The data generation process makes use of the learned gradient information and the efficient Langevin sampling method. It could be particularly useful for synthesizing minority class in imbalanced learning scenario. The generated samples are representative, credible and well-behaved: they interpolate and extrapolate existing data, mimic the behaviour of the underlying distribution (e.g. Fig.\ref{fig:new_sample_generation} and Fig.\ref{fig:imbalanced_classification_compare_new_data}). Data generation is made successful given a well-learned score function and proper sampling settings.

\paragraph{Generative classifier for imbalanced data}

The generative approach is principled and explainable, particularly in classifying imbalanced data: it acknowledges the fact of imbalance by encoding the imbalance ratio into priors. Therefore, it may be the natural way to model imbalance data. With score-based modelling, the task of generative classification becomes learning class-conditional score functions, construct class-conditional densities and compare posteriors. Scores are also involved in solving the decision boundary equation via \textit{Newton-Raphson} method (e.g. Fig.\ref{fig:Gaussian_pdfs_recovered_by_scores}). If additional distributional assumption is added, we may be able to coin the learned score functions and densities in analytical form (Fig.\ref{fig:Gaussian_pdfs_recovered_by_scores_assumeGaussian}). 

\paragraph{Tuning the (hyper)parameters}

It remains as an open question how to choose a proper neural architecture and its hyper-parameters (e.g. learning rate) to sufficiently, and not overly, extract gradient information from data. Shallow or deep, narrow or wide, high or low learning rates, \textit{etc}, could lead to underfitting or overfitting behaviours; they are general topics rooted in deep learning. We acknowledge that different networks can approximate the same score function with different (computational) complexities and accuracies. The image of neural architecture search is beyond this paper.

As an initial density value $\hat{p}_\theta(x_0)$ is required for score-based density construction, challenge lies in how to supply an educated guess of $\hat{p}_\theta(x_0)$ for each class. If the data is reasonably Gaussian clustered, we can use the Gaussian central probability $p(\hat{\mu)} = 1/(2\pi)^{d/2}|\hat{\Sigma}|^{1/2}$. However, misuse of the Gaussian guess could induce error when data distribution is not Gaussian (e.g. the fraud detection data) and if it's multi-modal (we might be supplying an badly estimated, non-representative point). In such case we have the purely empirical neighbourhood counting method.

There are also hyper-parameters in sampling. For example, in Langevin dynamics we have step size $\epsilon$, chain length $l$ and discard rate $\gamma$. The aim of sampling is to generate samples that follow the underlying distribution. To achieve this we perturb existing samples and produce new samples following the learned gradient fields, using proper hyper-parameter settings. However, we may still have problems of slow mixing of Langevin dynamics and ineffective learning in data scarcity regimes \cite{SM_Yang}.

\section{Future work}

This work builds some basic blocks and serves as an intro to discriminative and generative classification with score-based generative modelling. Future work may include extending the current framework to many other applications where learning-based data generation is demanded (e.g. rare event modelling, adversarial training), or density estimation is difficult (e.g. high dimensions). Gradient learning has the advantage that the learning outcome (i.e. the score function) is not constrained by a unit sum, and it can be represented by any input-output mapping (e.g. a parametric formula or an architecture). We could, for example, look into other task-specific learning representations for score functions, directly learning the score function of a discriminative density, making decisions by contrasting individual and global scores, or using scores as features in other machine learning tasks. Opportunities also exist in exploring other gradient-based sampling routines (e.g. HMC), utilising scores as an approximation; or comparing other generative modelling methods. Some automation process may be devised to train a proper score architecture to match a desired density profile, and tune the hyper-parameters in the sampling procedure. Generative modelling in general could be a cheap surrogate for active learning wherever data labelling is expensive. Although this work focuses on binary classification tasks, it can be extended to multi-class classification problems, using e.g. \textit{one-vs-one} or \textit{one-vs-all} contrasting strategies \cite{Intro_James}.

\section[title]{Related work\footnote{The related work list is by no means complete due to page limit.}}

On general classification, Friedman et al. \cite{elements_Friedman}, James et al. \cite{Intro_James}, Murphy \cite{ML_Murphy} and Berger \cite{decision_theory_Berger} gives comprehensive treatments of the classification decision theory. Ng and Jordan \cite{NIPS2001_Ng} compared logistic regression and naive Bayes classifiers, and showed that while discriminative learning has lower asymptotic error, a generative classifier may approach its (higher) asymptotic error much faster. Murphy \cite{ML_Murphy} described discriminative and generative classifiers in various settings. Rasmussen \cite{GP_Rasmussen} described the decision theory for both types of classification in Bayesian context. James et al. \cite{Intro_James} also have a treatise on different classifiers. For classification in the imbalanced domain, Chawla et al. \cite{SMOTE_Chawla} proposed the SMOTE oversampling approach for classifying imbalanced datasets and showed that a combination of oversampling and undersampling can achieve better classifier performance (measured by ROC) using C4.5, Ripper and a Naive Bayes classifiers. Moniz et al. \cite{SMOTEBoost_Moniz} proposed variants of SMOTEBoost which combine boosting technique and SMOTE resampling, and demonstratesd its use in imbalanced regression tasks such as extreme values prediction. He et al. \cite{ADASYN_He} presented the ADASYN approach for generating samples based on their level of difficulty in learning, which reduces the bias introduced by class imbalance and adaptively shifts the decision boundary. Pozzolo et al. \cite{credit_card_Pozzolo} argued that the bias due to undersampling significantly impacts classification accuracy and probability calibration, and tested the argument on the credit card transaction dataset which is also used in this work.

On sampling methods, Metropolis et al. \cite{Metropolis_Metropolis} proposed the the \textit{Metropolis algorithm}, Hastings \cite{MC_Hastings} introduced the Metropolis-Hastings sampler, Geman \cite{Gibbs_Geman} developed the Gibbs sampler, Kloek et al. \cite{importance_sampling_Kloek} introduced importance sampling, Simon et al. \cite{HMC_Duane} proposed the Hamiltonian Monte Carlo (HMC) algorithm, which mixes molecular dynamics and Langevin to guide MC simulation. In recent decades, sequential MC methods \cite{MC_Doucet,SMC_Chopin} are also advanced. While some MC methods may suffer from random walk behaviour, Welling and Teh \cite{Langevin_Welling} combined stochastic gradient optimization (with Langevin dynamics for noise injection) and Bayesian posterior sampling to enable efficient MCMC sampling and generate samples which converge to the full posterior distribution.

On score methods, Hyv\"arinen \cite{scoreMatching_Hyvarinen1} proposed the score matching method for estimating non-normalized statistical models, derived the simplified objective function, and validated it on multivariate Gaussian and independent component analysis models and image data. Hyv\"arinen \cite{scoreMatching_Hyvarinen2} extended score matching for binary variables and the non-negative real domain, and obtained in closed form for some exponential families. Yu et al. \cite{scoreMatching_Yu} described a generalized form of score matching for non-negative data with improved estimation efficiency, and improved theoretical guarantees of the regularized score matching method. Song and Ermon \cite{SM_Yang} proposed the combined use of score function and Langevin dynamics for generative sampling, and introduced perturbation in score estimation, which gives comparable performance to GANs. Song et al. \cite{SSM_Yang} proposed sliced score matching for complex models and higher dimensional data by projecting the scores onto random vectors, and applied it to deep energy-based models, variational inference and Wasserstein Auto-Encoders. Pacchiardi and Dutta \cite{scoreMatching_Lorenzo} used score matching for training a neural conditional exponential family to approximate the ABC likelihood, and applied it in MCMC sampling for intractable distributions and to large-dimensional time-series model. Generative models have been used as adversarially robust classifiers for complex datasets, particularly in the image classification domain \cite{score_classifiers_Yang}. Zimmermann et al. \cite{score_classifiers_Yang} investigated score-based generative classification of natural images, and found marginal advantage over discriminative classifiers in terms of adversarial robustness, yet it provides a different approach to classification.

\section{Conclusions}

Score-based generative method is efficient in terms of learning and sampling, robust to perturbations, effective in high dimensions and imbalanced situations. With sample-based score matching, learning complex (e.g. multi-modal) score functions are enabled by modern deep learning techniques. Sampling from a score function is convenient via Langevin dynamics; the sampling process is flexible, yields better distributional properties compared to interpolation methods, and is particularly useful in the presence of small, sparse, or disturbed data. Score-based generative classification, with comparable performance and marginal advantage, provides a novel method to classification; discriminative classification with score-generated data gives better performance over other data augmentation methods across metrics, as evidenced by simulated and real-world experiments.

\section{Code availability}

All codes are available on https://github.com/YongchaoHuang. 

\printbibliography

\begin{appendices}
\section{Derivation of discriminative densities} \label{App:score_of_discriminative_densities}

Here we provide details of deriving the score functions and gradients for the binary case in Section.\ref{Sec:score_assisted_discriminative_classification}. 
Using the specified representations of discriminative densities $p(y=0|x)=\frac{e^{-f_{\theta_0}(x)}}{Z_{\theta}}$ and $p(y=1|x)=\frac{e^{-f_{\theta_1}(x)}}{Z_{\theta}}$, where $Z_\theta(x)=e^{-f_{\theta_0}(x)} + e^{-f_{\theta_1}(x)}$, we have: 

\begin{multline} \label{Eq:discriminative_derivation_s0}
    s_{\theta_0}(x) = \nabla_x \log p(y=0|x) = \frac{\partial [-f_{\theta_0}(x) - \log Z_{\theta}]}{\partial x} = -f'_{\theta_0}(x) - \frac{1}{Z_\theta} Z'_\theta \\
    = -f'_{\theta_0}(x) - \frac{-f'_{\theta_0}(x) e^{-f_{\theta_0}(x)}-f'_{\theta_1}(x) e^{-f_{\theta_1}(x)}}{e^{-f_{\theta_0}(x)} + e^{-f_{\theta_1}(x)}} = \frac{[f'_{\theta_1}(x) - f'_{\theta_0}(x)]e^{-f_{\theta_1}(x)}}{e^{-f_{\theta_0}(x)} + e^{-f_{\theta_1}(x)}}
\end{multline}

\noindent where $f'(x)$ denotes derivative \textit{w.r.t.} $x$. Similarly, $s_{\theta_1}$ can be derived as: 

\begin{equation} \label{Eq:discriminative_derivation_s1}
    s_{\theta_1}(x) = \frac{[f'_{\theta_0}(x) - f'_{\theta_1}(x)]e^{-f_{\theta_0}(x)}}{e^{-f_{\theta_0}(x)} + e^{-f_{\theta_1}(x)}}
\end{equation}

\noindent We see that:

\begin{equation}
    s_{\theta_0}(x) + s_{\theta_0}(x) = \frac{[f'_{\theta_0}(x) - f'_{\theta_1}(x)][e^{-f_{\theta_0}(x)} - e^{-f_{\theta_1}(x)}]}{e^{-f_{\theta_0}(x)} + e^{-f_{\theta_1}(x)}}
\end{equation}

\begin{equation}
    s_{\theta_0}(x) - s_{\theta_0}(x) = f'_{\theta_1}(x) - f'_{\theta_0}(x)
\end{equation}

\begin{equation}
    \frac{s_{\theta_0}(x)}{s_{\theta_1}(x)} = -e^{f_{\theta_0}(x) - f_{\theta_1}(x)}
\end{equation}

\noindent For logistic densities, we have $f_{\theta_0}(x) = \theta^Tx$, $f_{\theta_1}(x) = 0$ and $Z_{\theta}=1+e^{-\theta^Tx}$.

\begin{equation}
    s_{\theta_0} = -\frac{\theta'}{1+e^{-\theta^Tx}}, 
    s_{\theta_1} = \frac{\theta' e^{-\theta^Tx}}{1+e^{-\theta^Tx}}
\end{equation}

\noindent and 

\begin{equation}
    s_{\theta_0}(x) + s_{\theta_0}(x) = - \frac{\theta' (1-e^{-\theta^Tx})}{1+e^{-\theta^Tx}},
    s_{\theta_0}(x) - s_{\theta_0}(x) = - \theta', 
    \frac{s_{\theta_0}(x)}{s_{\theta_1}(x)}= - e^{\theta^Tx}
\end{equation}

\noindent where $\theta'$ equals $\theta$ but with intercept removed after differentiation. 

As a comparison to score function, the gradient of the densities are:

\begin{equation} \label{Eq:binary_density_score_func_0}
    \nabla_x p_{\theta_0}(x) = \frac{\nabla_x f_{\theta_1}(x)-\nabla_x f_{\theta_0}(x)}{[1+e^{f_{\theta_0}(x)-f_{\theta_1}(x)}][1+e^{f_{\theta_1}(x)-f_{\theta_0}(x)}]}
\end{equation}
\noindent and $\nabla_x p_{\theta_1}(x) = -  \nabla_x p_{\theta_0}(x)$, as $p(y=1|x)=1-p(y=0|x)$ is satisfied everywhere. Logistic density, for example, gives:

\begin{equation} \label{Eq:logistic_density_derivative}
     \nabla_x p_{\theta_0}(x) = - \frac{\theta}{(1+e^{\theta^Tx})(1+e^{-\theta^Tx})}
\end{equation}

The scores and gradients of the learned logistic densities are compared in the simulated 1D and 2D Gaussian data cases in Fig.\ref{fig:1D_Gaussian_logistic_regression} and Fig.\ref{fig:2D_Gaussian_logistic_regression}, respectively.

\section{Gaussian scores and separality} \label{App:Gaussian_scores_and_separality}

Extending the discussion in Section.\ref{Sec:score-based generative classification}, here we take a closer look at the binary scenario where the two classes data are Gaussian distributed. We already know its score function $s(x|\mu,\Sigma)=\Sigma^{-1}(\mu-x)$. Following the generative decision rule (Eq.\ref{Eq:generative_classification_2}), for a test point $x=(x_1,x_2)$, a naive classifier would associate it with the class with higher density value, maybe with a soft margin $\gamma_0$ introduced such that $\hat{y}=0$ if $p_{\theta_1}(x)-p_{\theta_0}(x) \geq \gamma_0$. Here for simplicity we just use $\gamma_0=0$. The decision boundary is then determined by $p_{\theta_0}(x^*)=p_{\theta_1}(x^*)$ where we denote the points on the decision boundary as $x^*$.

Substituting the multivariate Gaussian density (Eq.\ref{Eq:MND_pdf}) into the the equal density condition, we arrive at the boundary equation (BE): 
\begin{multline} \label{Eq:Gaussian_decision_boundary_1}
    BE(x^*) = \log p_{\theta_1}(x^*) - \log p_{\theta_0}(x^*) \\
    = -\frac{1}{2} x^T(\Sigma_1^{-1}-\Sigma_0^{-1})x + (\mu_1^T\Sigma_1^{-1} - \mu_0^T\Sigma_0^{-1})x \\+ \frac{1}{2} \log (| \Sigma_0 |/| \Sigma_1 |) + \frac{1}{2}(\mu_0^T \Sigma_0^{-1} \mu_0 - \mu_1^T \Sigma_1^{-1} \mu_1) = 0
\end{multline}

\noindent where $|\cdot|$ denotes matrix determinant. This quadratic boundary curve or surface is also used in QDA \cite{Intro_James}. The first-order derivative of $BE(x)$:

\begin{equation} \label{Eq:boundary_equation_Gaussian_derivative}
    \frac{\partial BE(x)}{\partial x} = (\Sigma_0^{-1} - \Sigma_1^{-1})x + \Sigma_1^{-1}\mu_1 - \Sigma_0^{-1}\mu_0
\end{equation}

\noindent Again, Eq.\ref{Eq:boundary_equation_Gaussian_derivative} can be used to locate the roots of Eq.\ref{Eq:Gaussian_decision_boundary_1} in a \textit{Newton-Raphson} scheme. The boundary between two one-dimensional Gaussian densities issued by Eq.\ref{Eq:Gaussian_decision_boundary_1}, for example, is: 

\begin{multline}
    (1/\sigma_1^2-1/\sigma_0^2)x^2 - 2(\mu_1/\sigma_1^2 - \mu_0/\sigma_0^2)x - \log (\sigma_0^2/\sigma_1^2) - (\mu_0^2/\sigma_0^2- \mu_1^2/\sigma_1^2) = 0
\end{multline}

\noindent with the discriminative points $x=(\mu_1\sigma_0-\mu_0\sigma_1)/(\sigma_0-\sigma_1)$ and $x=(\mu_1\sigma_0+\mu_0\sigma_1)/(\sigma_0+\sigma_1)$, assuming $\mu_1 \geq \mu_0$. 

Further, if the two Gaussian classes share the same covariance, i.e. $\Sigma_1 = \Sigma_2= \Sigma$, Eq.\ref{Eq:Gaussian_decision_boundary_1} reduces to:

\begin{equation} \label{Eq:Gaussian_decision_boundary_2}
    (\mu_1^T - \mu_0^T) \Sigma^{-1} x^* = \frac{1}{2} (\mu_1^T \Sigma^{-1} \mu_1 - \mu_0^T \Sigma^{-1}\mu_0)
\end{equation}

\noindent which is a linear boundary. This is the disciminator for LDA \cite{Intro_James}. In one-dimensional case, two Gaussians with same standard deviation will meet at the equal probability point $x^*=(\mu_1+\mu_2)/2$. 

As with this naive classifier, for each class, all points located beyond $x^*$ are mis-labelled. The mis-classification rate is controlled by the probability that those points are from the beyond-boundary arena, which is: 

\begin{equation}
    p(x > x^*) = 1 - \Phi [L^{-1}(x^*-\mu)]
\end{equation}

\noindent here $x>x^*$ denotes that $x$ is beyond the boundary (with any coordinate exceeding the boundary), $\Phi(z)=\int_{-\infty}^z \mathcal{N}(x|\textit{\textbf{0}},\textit{\textbf{I}}) dx$ is the CDF of standard normal, and $L$ is the lower triangular matrix in covariance cholesky decomposition $\Sigma=LL^T$. The total probability of mis-classifying an unknown point $x$ is $p_{mis}(x)=p_0(x > x^*) + p_1(x > x^*)$, with only one term of the summation is non-zero at any time. 

If we have learned the score functions $s_{\theta_0}$ and $s_{\theta_1}$, we can also construct BE by integration. Here are describe the details of finding the boundary points starting from scores, as used before in constructing the score-based decision boundary in Fig.\ref{fig:2D_Gaussian_generative_boundary}. The first step is to recover the density ratio from score functions. As the mapping between score function and density is a differential equation, we can simply inverse it by integration. Then we apply the simple generative decision rule of equal density to find the boundary. We start by writing the (log) density ratio differential equation:
\begin{multline} \label{Eq:Gaussian_density_ratio}
    s_{\theta_1}(x) - s_{\theta_0}(x) = \nabla_x \log p_{\theta_1}(x) - \nabla_x \log p_{\theta_0}(x) = \nabla_x \log \frac{p_{\theta_1}(x)}{p_{\theta_0}(x)}
\end{multline}

\noindent By integration we can solve for the \textit{pdf} ratio function $R(x)$:

\begin{equation} \label{Eq:pdf_ratio_func1}
    R(x) = \frac{p_{\theta_1}(x)}{p_{\theta_0}(x)} = Ce^{\int_{\Omega} [s_{\theta_1}(x) - s_{\theta_0}(x)] dx}
\end{equation}

\noindent where the integration over the intersection support domain $\Omega$ can be finite or improper. Unfortunately, neither the differentiation (Eq.\ref{Eq:Gaussian_density_ratio}) nor integration (Eq.\ref{Eq:pdf_ratio_func1}) form informs any corresponding equality between densities and scores: if $s_{\theta_1}(x)=s_{\theta_0}(x)$, from Eq.\ref{Eq:Gaussian_density_ratio} we have $p_{\theta_1}(x) = C p_{\theta_0}(x)$, i.e. the two densities can off-set by a multiplicative constant; if $p_{\theta_1}(x)=p_{\theta_0}(x)$, we have the equation for a normal decision boundary $\int_\Omega [s_{\theta_1}(x) - s_{\theta_0}(x)] dx + \log C = 0$ (obtained by setting $R(x^*)=1$), which doesn't give any hints on the local (e.g. at particular points) equality between $s_{\theta_0}(x)$ and $s_{\theta_1}(x)$.

In practice, we are done because  we have the learned $s_{\theta_0}(x)$ and $s_{\theta_1}(x)$ from data (optimally in parametric form), and $C$ can be determined by empirically estimate an initial point density ratio $\hat{p}_{\theta_1}(x_0)/\hat{p}_{\theta_0}(x_0)$. They together can be substituted into $\int_\Omega [s_{\theta_1}(x) - s_{\theta_0}(x)] dx + \log C = 0$ to find $x^*$, which was used in Fig.\ref{fig:2D_Gaussian_generative_boundary}. However, given the analytical form of Gaussian densities, we shall be able to further derive an analytical formula of $C$, which can provides convenient sample-based estimate of $C$ without using an initial point to evaluate $\hat{p}_{\theta_1}(x_0)/\hat{p}_{\theta_0}(x_0)$. To this regard, we first substitute the analytical score function $s(x)=\Sigma^{-1}(\mu-x)$ into $R(x)$: 

\begin{equation}
    R(x)
    = Ce^{-\frac{1}{2} x^T(\Sigma_1^{-1}-\Sigma_0^{-1})x + (\mu_1^T\Sigma_1^{-1} - \mu_0^T\Sigma_0^{-1})x}
\end{equation}

\noindent Comparing the above formula with exact density ratio (referencing Eq.\ref{Eq:MND_pdf}) gives:

\begin{equation} \label{Eq:Gaussian_boundary_equation_constant}
    \log C= \frac{1}{2} \log (| \Sigma_0 |/| \Sigma_1 |) + \frac{1}{2}(\mu_0^T \Sigma_0^{-1} \mu_0 - \mu_1^T \Sigma_1^{-1} \mu_1)
\end{equation}

With Eq.\ref{Eq:Gaussian_boundary_equation_constant}, under the assumption of Gaussianalities, we can conveniently estimate the constant $C$ via Eq.\ref{Eq:Gaussian_boundary_equation_constant} using sample-based means and covariances.

\end{appendices}

\end{document}